\documentclass{article} 
\usepackage{iclr2024_conference,times}
\iclrfinalcopy
\usepackage{smile}

\usepackage{amsmath,amsfonts,bm}

\def\eqref#1{equation~\ref{#1}}

\def\1{\bm{1}}

\DeclareMathAlphabet{\mathsfit}{\encodingdefault}{\sfdefault}{m}{sl}
\SetMathAlphabet{\mathsfit}{bold}{\encodingdefault}{\sfdefault}{bx}{n}

\usepackage{amssymb,amsmath,amsthm}

\usepackage{url}
\usepackage{xcolor}         
\definecolor{green}{RGB}{38,128,96}
\usepackage{colortbl}
\usepackage{multirow}
\usepackage{caption}
\usepackage{subfigure}
\usepackage{enumitem}
\usepackage[colorlinks,citecolor=green]{hyperref}
\usepackage{wrapfig}
\usepackage{booktabs}
\usepackage{listings}
\lstdefinestyle{mystyle}{
  basicstyle=\ttfamily,
  frame=single,
  breaklines=true,
  breakindent=1pt,
  backgroundcolor=\color{blue!5}, 
}

\title{Explanation-aware Soft Ensemble Empowers Large Language Model In-context Learning}

\author{Yue Yu$^{\spadesuit}$\thanks{Work done during the internship at Google Research. E-mail: \texttt{yueyu@gatech.edu}}, Jiaming Shen$^{\clubsuit}$, Tianqi Liu$^{\clubsuit}$, Zhen Qin$^{\clubsuit}$, Jing Nathan Yan$^{\diamondsuit*}$, Jialu Liu$^{\clubsuit}$,  \\ 
\bf Chao Zhang$^{\spadesuit}$, Michael Bendersky$^{\clubsuit}$\\
$^{\spadesuit}$ Georgia Institute of Technology \quad
$^{\clubsuit}$ Google Research \quad
$^{\diamondsuit}$ Cornell University \\
}

\usepackage{xspace}

\newcommand{\method}{\textsc{EaSE}}
\newcommand{\weight}{explanation-aware ensemble\xspace}
\newcommand{\aggregation}{soft probability aggregation\xspace}
\newcommand{\mquote}[1]{{``\emph{#1}''}}

\newcommand{\blue}[1]{\textcolor{blue}{#1}}
\newcommand{\red}[1]{\textcolor{red}{#1}}

\begin{document}

\maketitle

\begin{abstract}
Large language models (LLMs) have shown remarkable capabilities in various natural language understanding tasks. 
With only a few demonstration examples, these LLMs can quickly adapt to target tasks without expensive gradient updates. 
Common strategies to boost such ``in-context'' learning ability are to ensemble multiple model decoded results and require the model to generate an explanation along with the prediction. 
However, these models often treat different class predictions equally and neglect the potential discrepancy between the explanations and predictions.
To fully unleash the power of explanations, we propose {\method}, an \emph{Explanation-Aware Soft Ensemble} framework to empower in-context learning with LLMs. 
We design two techniques, explanation-guided ensemble, and soft probability aggregation, to mitigate the effect of unreliable explanations and improve the consistency between explanations and final predictions. 
Experiments on seven natural language understanding tasks and four varying-size LLMs demonstrate the effectiveness of our proposed framework.
\end{abstract}

\section{Introduction}

Recent advancements in Natural Language Processing (NLP) have witnessed the remarkable capabilities of Large Language Models (LLMs)~\citep{brown2020language,tay2022ul2,chowdhery2022palm,palm2,touvron2023llama2,openai2023gpt4}.
These LLMs can rapidly adapt to new tasks by learning only on a few input-output pairs (\emph{a.k.a.} demonstrations) in context, without any gradient update~\citep{wei2022emergent,xie2022an}.
Yet, beyond those demonstrations, a significant facet of human learning revolves around explanations. 
These explanations\footnote{In this paper, we use the term `explanations' and `rationales' interchangeably.}, typically in the form of a few keywords or sentences, reveal the underlying principles connecting the input and output~\citep{zaidan2007using,narang2020wt5}.  
Consequently, the integration of free-text explanations into LLM prompting holds great potentials to further enhance in-context learning performance.

Recent studies have examined how to incorporate free-text explanations into LLM in-context learning scheme.
For instance, the \emph{Predict-then-Explain} pipeline \citep{lampinen2022language} proposes to generate the explanation \emph{after} making the prediction.
Consequently, the predictions from LLM won't directly benefit from their corresponding explanations.
In contrast, the \emph{Explain-then-Predict} pipeline (also called ``Chain-of-Thought'') \citep{nye2021show,wei2022chain} generates explanations \emph{before} making predictions via greedy sampling.
When the LLM-generated explanations are unreliable, predictions from this approach will be largely distracted and defective~\citep{ye2022unreliability}.
To mitigate this issue, \cite{wang2023selfconsistency} improves the ``Chain-of-Thought'' pipeline by first generating multiple predictions with different explanations using temperature sampling and then aggregating them via majority voting.
However, this approach can be sub-optimal as (1) temperature sampling increases the inconsistency between generated explanations and their associated class predictions, and (2) majority voting treats different predictions associated with explanations of varying qualities equally.
As a result, how to robustly leverage natural language explanations for empowering LLM in-context learning remains an open research question.

\begin{figure}[t]
    \centering
    \vspace{-2ex}
    \includegraphics[width=0.95\linewidth]{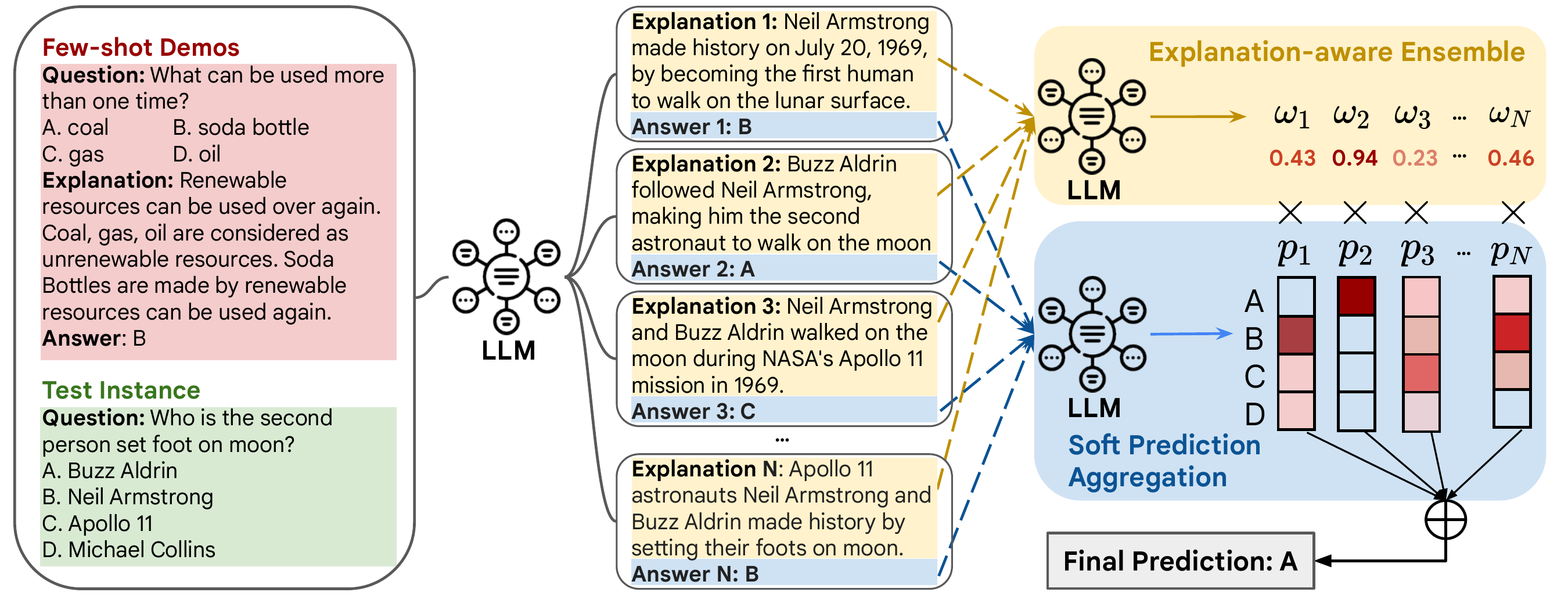}
    \vspace{-1ex}
    \caption{The overview of {\method} framework. 
    }
    \vspace{-0.5ex}
    \label{fig:overall}
\end{figure}

In this work, we present a novel {\underline{\textbf{E}}xplanation-\underline{\textbf{a}}ware \underline{\textbf{S}}oft \underline{\textbf{E}}nsemble} framework, named {\method}, to assist LLM in-context learning with explanations.
Our technique integrates explanations into the ensemble procedure and employs soft probability to mitigate discrepancies between explanations and predictions. 
The key module of the {\method} framework hinges upon the idea of weighted ensemble: As shown in Figure~\ref{fig:overall}, instead of treating all predictions equally, we assign a score to each prediction based on the contextual relevance and inherent quality of its associated explanation, which will be used as a weight during the final ensemble stage. 
This explanation-aware ensemble stage is also realized with an LLM --- after generating explanations and predictions using temperature sampling for each test instance, we prompt the LLM to weight all class predictions based on their associated explanations in an in-context manner. 
While the LLM offers great promise for the weighting purpose,  it is crucial to provide sufficient \emph{supervision signals} as demonstrations to guide the LLM scoring, yet the primary constraint for this step lies in the absence of \emph{negative} explanations from few-shot demonstrations. 
To construct negative examples efficiently, we first use LLM to generate explanations for few-shot demonstrations, then select explanations associated with \emph{incorrect predictions} as the negative samples.
In this way, the LLM scorer can be readily applied to perform explanation-aware ensembling without any additional annotation.

Beyond explanation-aware ensembling, {\method} incorporates an additional technique named \emph{soft probability aggregation}, 
which helps to mitigate the \emph{inconsistency} between explanations and predictions, given the sampling process may inevitably infuse noises into the final prediction. 
Specifically, it employs probabilities across various class-indicative verbalizers in place of the original one-hot predictions.
This design, although conceptually simple, can effectively reduce the discrepancies between explanations and predictions and further improve the final predictions accuracy.

Our contributions can be summarized as follows:
\begin{itemize}[leftmargin=0.5cm]
    \item We propose the \method{} framework to better facilitate in-context learning for large language models with natural language explanations. 
    \item We design two techniques, namely \weight{} and \aggregation{}, to enable the model to focus on predictions associated with explanations of higher qualities while reducing the inconsistency between explanations and predictions. 
    \item We conduct experiments on seven natural language understanding (NLU) datasets spanning between natural language inference (NLI) and question answering (QA), and our method outperforms previous state-of-the-art approaches using different LLMs as the backbone.  Our analysis further justifies the advantages of using LLMs for explanation weighting to support correct answer candidates and leveraging \aggregation{} to mitigate inconsistent predictions.

\end{itemize}



\section{Related Work}
\label{sec:related}
Two prevalent explanation types exist for interpreting NLP models: (1) \emph{extraction-based explanations} that highlight important segments of the original input text~\citep{zhang2016rationale,eraser,paranjape-etal-2020-information,zhou2020towards,yin2022interpreting} and (2) \emph{free-form explanations} that craft prediction rationales directly using natural language text~\citep{rajani-etal-2019-explain,sun2022investigating,wiegreffe-etal-2021-measuring,wiegreffe-etal-2022-reframing,wang2023pinto,ludan2023explanation}.
Beyond aiding in model interpretation, recent studies have demonstrated that these explanations can also enhance the few-shot learning capabilities of large language models.
For example, \cite{wei2022chain,zelikman2022star} propose to \emph{prepend explanations} before the answers while \cite{lampinen2022language} suggest adding \emph{post-answer explanations}.
Given that these explanations are often derived during the LLM decoding stage and may contain noise, \citep{wang2023selfconsistency,wang2022rationale} advocate for generating multiple candidate explanations with their respective predictions, followed by aggregating these predictions via majority voting.
In our study, we focus on \emph{free-form explanations} and explore how to better aggregating these predictions with explanations in a weighted ensemble.
Using a bootstrapped LLM, we subsequently evaluate each explanation to enhance in-context learning outcomes.
Another line of research related to our study is automated explanation quality evaluation~\citep{sun2022investigating,joshi2023machine,wiegreffe-etal-2021-measuring,chen2023rev,chen2023models}.
\cite{ye2022unreliability} utilize lexical features to measure the faithfulness of explanations without considering their semantics.
\cite{chen2021can,li2023making} leverage a NLI fine-tuned model to verify the explanations reliability.
\citep{fu2023gptscore,liu2023gpteval,qin2023large,chen2023alpagasus} also study how to use LLM to build a generic text quality scorers for generation and ranking tasks. 
These studies often rely on additional ground-truth labels and human annotations, making them less suitable when the labels for test instances are unknown. 
In contrast, our research diverges from the pure evaluative perspective while focusing more on effectively leveraging model-generated explanations to empower the LLM  in-context learning performance. 
There are also several works that attempted to use LLMs to generate demonstrations~\citep{shao2023synthetic,kim2023aligning,yu2023large}, but they mainly focus on producing few-shot demonstrations, whereas our approach emphasizes the generation of negative examples for more robust scoring and evaluation of explanations.







\section{Method}
In this section, we first give a brief introduction to the problem definition. Then, we present our approach with two designs, namely \weight{} and \aggregation{}, with the goal of leveraging the generated explanations to improve the final prediction performance.
\subsection{Problem Definition}
In this task, we are given a LLM $\cM$ parameterized by $\theta$,  a set of few-shot demonstrations $\cD=\{(x_i, e_i, y_i)\}_{i=1}^K$ on a target classification task\footnote{Future work would be suited to consider extending our work to generative tasks.}, where $K$ is the number of demonstrations, $x_i, y_i$ are the input text and label for the $i$-th example, and $e_i$ is the corresponding ground-truth explanation. 
For each test example $x \in \cD_{\text{test}}$, we aim to leverage $\cM$ and $\cD$ to predict its own label. 
Our primary goal is to improve the prediction accuracy for test examples.

\subsection{Recap of Self-consistency Pipeline for In-context Learning}
Here we give a brief introduction to the self-consistency  approach \citep{wang2023selfconsistency}. For each test example $x\in\cD_{\text{test}}$, it first forms the prompt for few-shot demonstrations as   
$\mathcal{P}=\left\{\cT, \operatorname{shuffle}(\|_{i=1}^K(x_i, e_i, y_i)\right)\}$, where $\cT$ is the prompt template, and $ \operatorname{shuffle}\left(\|_{i=1}^K(x_i, e_i, y_i)\right)$ is a permutation of $K$ demonstrations. 
Then, it generates $N$ candidate explanations together with predictions (denoted as $(e_j, p_j)$) via sampling from the LLM with non-zero temperature as 
\begin{equation}
\small
    (e_j, p_j)_{j=1}^{N} \sim p_{\theta}\left(e, p \mid \cP, x\right),
    \label{eq:ep}
\end{equation}
Finally, it aggregates these $N$ candidates into the final prediction  via majority voting as
\begin{equation}
\small
    \tilde{y} = \argmax_{y}~\sum_{j=1}^{N}\mathbb{I}(p_j = y).
\end{equation}
Self-consistency enhances the standard explain-then-predict pipeline by utilizing multiple predictions derived from varied explanations. 
Despite its strong performance, through our examination, we've pinpointed two primary bottlenecks within the self-consistency pipeline, listed as follows:
\begin{itemize}[leftmargin=0.6cm]
    \item \emph{Explanation-agnostic Ensembling}: 
     Self-consistency uniformly weights all predictions and aggregates them via simple majority voting. This approach overlooks the variance in explanation quality, which can be problematic when certain predictions stem from flawed reasoning paths evident in poor-quality explanations.
    \item \emph{Explanation-Prediction Inconsistency}: During its prediction phase,  Self-consistency  employs the temperature sampling technique to draw samples from the LLM. This sampling step can introduce noise, leading to predictions that are inconsistent with their corresponding explanations~\citep{ye2022unreliability}.
\end{itemize}

The identified limitations necessitate the need for new techniques to better harvest intermediate explanations for obtaining the final prediction.  
Towards this goal, we propose our framework \method{}, which is tailored to tackle the aforementioned challenges. 
\method{} is comprised with two techniques, \weight{} and \aggregation{}, to optimize the LLM's prediction accuracy when deriving final outcomes from multiple candidate explanations.





\subsection{Explanation-guided Ensemble}


LLMs typically produce multiple explanations along with their predictions through a sampling process. 
Due to the intrinsic randomness of this sampling, the quality of these predictions can fluctuate.
To address the potential pitfalls where erroneous explanations results in inaccurate predictions, we introduce the \emph{\weight{}} technique.
This method estimates the significance of each class prediction based on its corresponding explanation. 
Consequently, our \weight{} technique ensures that predictions linked with better explanations carry greater weight during the final prediction aggregation phase.


\textbf{LLM as Explanation Scorer}
To evaluate various explanations, past research has either measured the lexical overlap between the explanation and the input text~\citep{ye2022unreliability} or employed models fine-tuned for NLI tasks~\citep{chen2021can,li2023making}.
In contrast to these methods, which either overlook the deep semantics of explanations or require extra human-annotated data, our explanation scorer is developed based on the powerful LLM $\cM$, directly harnessing its inherent linguistic and reasoning capabilities.

Given the original task input $x$ and one explanation $e$, we use the verbalizer $v_{\text{pos}} (v_{\text{neg}}$) to represent the class of this explanation being \mquote{positive} (\mquote{negative}).
A \mquote{positive} explanation means this explanation can help the model reach correct answer and a \mquote{negative} explanation means the other way around.
Then, we craft a supplementary prompt $\cT_{\text{score}}=$  \mquote{Can this explanation be used to help the model answer the question?} for LLM prompting.  
With the verbalizers and prompts, we effectively recast the problem of explanation scoring into determining the conditional probability of producing the verbalizer aligned with the positive label  $v_{\text{pos}}$, expressed as
\begin{equation}
\small
    \omega_e = p_{\theta}\left(y= v_{\text{pos}}\mid \cT_{\text{score}}, x, e\right).
\label{eq:llm_scorer}
\end{equation}
In this way, the score $\omega_e$ is normalized between 0 and 1 and a higher score indicates the explanation with better quality.

\begin{wrapfigure}[16]{r}{0.59\textwidth}
  \centering
  \vspace{-2ex}
  \includegraphics[width=0.60\textwidth]{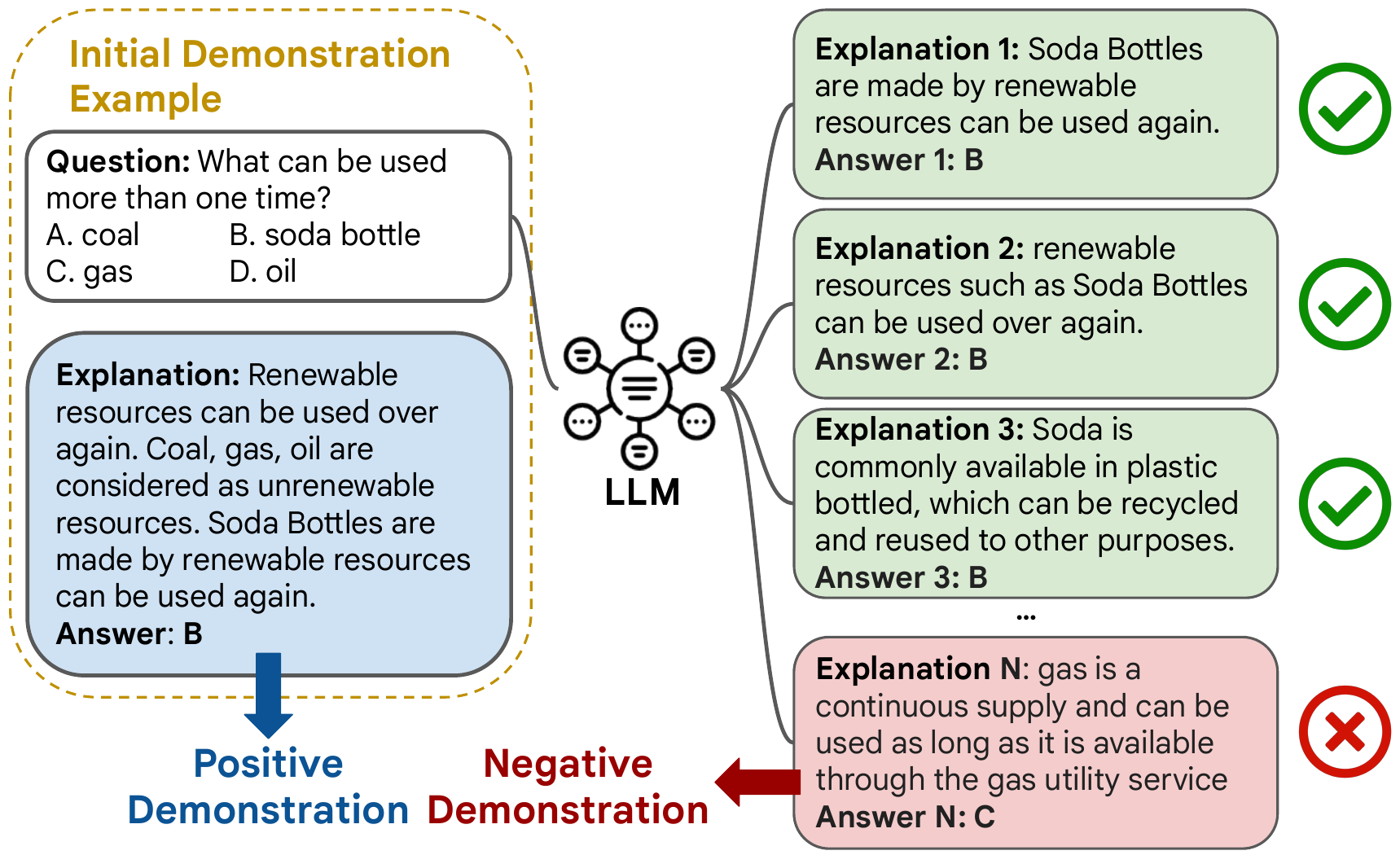}
  \caption{Bootstrapped LLM Scorer.}
  \vspace{-3ex}
\end{wrapfigure}

\textbf{Bootstrapped LLM Scorer} Although the above approach can already produce scores for each prediction, the score generated with the LLM $\cM$ can still be biased and less precise~\citep{wang2023large}, especially under the zero-shot scenario where no demonstrations are provided. 
To mitigate the bias and generate reliable scores, we propose to provide additional examples to serve as \mquote{positive} and \mquote{negative} explanations to facilitate LLM scoring using the original few-shot demonstrations in $\cD$.

For each original demonstration instance, it is straightforward to obtain  \mquote{positive} examples from the ground-truth explanation. Obtaining \mquote{negative} examples, on the other hand, can be more challenging as they are not explicitly provided. 
To tackle this issue, we exploit the assumption based on the utility of explanations: an ideal explanation should guide the model towards the accurate prediction of ground-truth labels~\citep{wiegreffe-etal-2021-measuring}. Consequently, it's reasonable to classify explanations leading to erroneous predictions as "negative". 
In practice, for every instance  $(x_i, y_i) \in \cD$, we randomly select $k$ (8 in this work) exemplars from the training set and then use these as demonstrations and generate a set of candidate pairs $\cC_i = \{(e_{ij}, p_{ij})\}_{j=1}^{N}$ via sampling from the LLM.  
Then, if the explanation-prediction pair $(e_{ij}, p_{ij})$ from $\cC_i$ satisfies $y_i \neq p_{ij}$, we select $e_{ij}$ to serve as the {negative} explanation set $\cN_i$ for $x_i$ as 
\begin{equation}
\small
\setlength{\abovedisplayskip}{5pt}
\setlength{\belowdisplayskip}{5pt}
   \cN_i=\{(e_{ij}, p_{ij}) \in \cC_i \mid y_i \neq p_{ij}\}. 
\end{equation}
To finalize the demonstration set for the LLM scoring step, we balance between  \mquote{positive} and  \mquote{negative} explanations: only instances possessing negative explanations (i.e. with non-empty $\cN_i$) are incorporated into the demonstrations; 
For every instance, a single negative explanation is chosen at random from the respective candidate set. 
This methodology produces a balanced demonstration set for LLM-based explanation scoring without requiring extra human annotations.


\subsection{Soft Probability Aggregation}
In the preceding step, the primary objective is to assign a score to each prediction based on its associated explanation through the LLM $\cM$. 
This process, however, does not account for directly modeling the LLM's output predictions. 
To bridge this gap, we propose \emph{soft probability aggregation}, a simple and intuitive approach to resolve the discrepancy between the explanations and predictions --- 
rather than aggregating over the raw predictions, it directly computes the  sum of the probabilities associated with each potential label, expressed as
\begin{equation}
\small
\setlength{\abovedisplayskip}{5pt}
\setlength{\belowdisplayskip}{5pt}
    \tilde{y} = \argmax_{y}
    ~\sum_{j=1}^{N} p_{\theta}\left(y \mid \cP, x, e_j\right).
\label{eq:soft_prob}
\end{equation}
The \emph{soft probability aggregation} addresses the noise inherited in different LLM sampling-based decoding algorithms, resulting in a more accurate and refined final prediction.


\subsection{Summary}
By plugging these two techniques together, we obtain the final prediction $\tilde{y}$ for the test instance $x$ as
\begin{equation}
\small
\setlength{\abovedisplayskip}{5pt}
\setlength{\belowdisplayskip}{5pt}
    \tilde{y} = \argmax_{y}~\sum_{j=1}^{N} \omega_{e_j} \times p_{\theta}\left(y \mid \cP, x, e_j\right),
\label{eq:overall}
\end{equation}
where $e_j$ is the intermediate explanations generated via Eq.~\ref{eq:ep}, the $\omega_{e_j}$ is the weight for  $e_j$ using the bootstrapped LLM scorer using Eq.~\ref{eq:llm_scorer}, and $p_{\theta}\left(y \mid \cP, x, e_j\right)$ is the soft probability generated using Eq.~\ref{eq:soft_prob}. 
Overall, calculating the score for each explanation and the soft probability both take an additional $O(N)$ time complexity. Fortunately, these two steps do not require additional model training and can be efficiently supported with distributed inference techniques in practice.
Other than these two techniques, our framework keeps other components intact and can be plugged into most LLM backbones for empowering its in-context learning ability.

\section{Experiments}

\subsection{Experiment Setups}
\textbf{Tasks}
We evaluate our \method{} framework on two types of tasks: natural language inference and question answering. 
Specifically, we use the following datasets:
(1) \textbf{E-SNLI} \citep{camburu2018snli}  is an enriched version of the Stanford Natural Language Inference (SNLI) corpus~\citep{bowman2015large}, augmented with human-annotated natural language explanations for entailment relations;
(2) \textbf{ANLI-R1/R2/R3} \citep{anli} is a set of three collections of adversarially generated NLI examples curated through a human-in-the-loop process;
(3) \textbf{ECQA} \citep{ecqa} is built upon CommonsenseQA benchmark~\citep{commonsenseqa} and contains additional human-annotated question explanations; 
(4) \textbf{OpenbookQA} \citep{openbookqa} is a QA dataset that requires comprehensive understanding and reasoning from open-book sources. As no ground-truth explanations are given, we use the provided facts for each question as the proxy explanations. 
(5) \textbf{StrategyQA} \citep{geva2021did}  focuses on reasoning over complex, multi-hop questions that often require strategic planning and decision-making.

\textbf{Baselines} We consider the following baselines:
(1) \textbf{Standard In-context Learning (ICL)} \citep{brown2020language}: it solely uses the input-label pairs for few-shot learning without using natural language explanations.
(2) \textbf{Predict-then-Explain (PE)}~\citep{lampinen2022language}: it provides the explanation after the labels for each instance when constructing prompts for demonstrations. During the inference stage, it generates the explanation after the prediction.  
(3) \textbf{Explain-then-Predict (EP)}~\citep{wei2022chain}: it is the standard chain-of-thought pipeline which provides an explanation before the label for demonstrations. During the inference stage, it first generates an explanation, then followed by the prediction. Note that for both PE and EP method, we use greedy sampling to obtain the explanation and prediction.
(4) \textbf{Self-consistency}~\citep{wang2022rationale,wang2023selfconsistency}: it improves over the standard EP pipeline by aggregating over multiple explanations from LLMs to enhance the robustness of the results.
(5) \textbf{FLamE} \citep{zhou2023flame} is a recent LLM few-shot learning method that generates multiple label-conditioned explanations and determines the final prediction based on the label that achieves the highest logit after reviewing all explanations for the given instance\footnote{In the original FLamE paper, the RoBERTa is used for final classification. For a fair comparison, we adjusted FLamE to use the in-context LLM as the classifier.}.

\paragraph{Implementation Details}
In our main experiments, we use PaLM2-S and PaLM2-L \citep{palm2} as the  backbone model.
Results on more (open source) backbone models are reported in Section~\ref{sec:open_source}.
For each dataset, we set the size of few-shot examples to 48 following \citep{zhou2023flame,marasovic2022few}, and fit as many instances as possible during inference until reached the maximum length. 
As the LLM is often sensitive to the selection of few-shot examples~\citep{yu2023cold,ye2023explanation,liu2022makes}, for each dataset, we create 5 splits from the original dataset, each containing 300 test examples, and report the average performance over 5 splits. During sampling, we set the default temperate to $t=0.7$ and sample $N=9$  candidate explanations for each instance.

\subsection{Overall Results}
\begin{table}[tbp]
  \label{tab:datasets}
  \renewcommand\arraystretch{0.95}
  \caption{The main experiments results, where ``BLS'' stands for bootstrapped LLM scorer and ``SPA'' stands for soft probability aggregation.
  }
  \resizebox{0.98\linewidth}{!}{
  \begin{tabular}{lccccccccc}
  \toprule
  \bfseries Backbone & \bfseries Methods & \bfseries E-SNLI & \bfseries ANLI-R1 & \bfseries ANLI-R2 & \bfseries ANLI-R3 &  \bfseries ECQA &\bfseries StrategyQA & \bfseries OpenbookQA & \bfseries Average \\
  \midrule
  \multirow{8}{*}{PaLM 2-S} 
  & ICL\scriptsize{~\citep{brown2020language}} & 59.88 & 54.38 & 48.10 & 52.66 & 59.84 & 66.69 & 80.21 & 60.25  \\
  & PE\scriptsize{~\citep{lampinen2022language}} & 71.02 & 62.59 & 55.18 & 57.17 & 74.39 & 71.75 & 79.70 & 67.40  \\
  & EP\scriptsize{~\citep{wei2022chain}} & 64.53 & 57.40 & 53.00 & 53.33 & 72.11 & 72.40 & 81.38 & 64.88 \\
  & Self-consistency\scriptsize{~\citep{wang2023selfconsistency}} & 68.68 & 65.40 & 56.49 & 59.00 & 74.48 & 76.94  & 83.47 & 69.21 \\
  & FLamE\scriptsize{~\citep{zhou2023flame}} & 67.58 & 60.36 & 52.00 & 50.15 & 72.80 & 75.33 & 80.14 & 65.48  \\
   \cmidrule(lr){2-10}  
  &\cellcolor{green!15} \method{} & \cellcolor{green!15}\bf 75.01 
  & \cellcolor{green!15}66.48 
  & \cellcolor{green!15}\bf 59.66 
  & \cellcolor{green!15}\bf 64.33 
  & \cellcolor{green!15}\bf 75.59 
  & \cellcolor{green!15}78.23 
  & \cellcolor{green!15}\bf 84.10 
  & \cellcolor{green!15}\bf 71.92 ($\uparrow$3.91\%)\\
  & \method{} w/o BLS & 73.84 & \bf 66.84 & 58.74 & 62.66 & 75.17 & \bf 78.40 & 83.91 & 71.37 \\
  & \method{} w/o SPA & 69.82 & 67.77 & 58.50 & 62.50 & 75.42 & 78.33 & 83.68 & 70.73 \\
  \midrule
  \multirow{8}{*}{PaLM 2-L} & 
  ICL\scriptsize{~\citep{brown2020language}} & 87.42 & 79.00 & 68.33 & 65.65 & 81.29 & 81.13 & 91.17 & 79.14 \\
  & PE\scriptsize{~\citep{lampinen2022language}} &  88.84 & 80.55 & 71.49 & 68.33 & 83.13 & 83.19 & 92.46 & 81.14  \\
  & EP\scriptsize{~\citep{wei2022chain}} &  84.59 & 79.03 & 67.99 & 67.66 & 80.51 & 85.45 & 89.74 & 79.28 \\
  & Self-consistency\scriptsize{~\citep{wang2023selfconsistency}} & 87.34 & 81.29 & 73.16 & 70.16 & 82.67 & 87.85 & 92.88 & 82.19  \\
  & FLamE\scriptsize{~\citep{zhou2023flame}} & 83.23 & 71.85 & 58.50 & 56.83 & 80.26 & 84.79 & 93.14 & 75.51 \\
   \cmidrule(lr){2-10}  
  & \cellcolor{green!15}\method{} 
  & \cellcolor{green!15}\bf 89.42 
  & \cellcolor{green!15}\bf 83.69 
  & \cellcolor{green!15}\bf 76.16 
  & \cellcolor{green!15}\bf 74.00 
  & \cellcolor{green!15}\bf 83.65 
  & \cellcolor{green!15}\bf 89.90 
  & \cellcolor{green!15}\bf 93.93 
  & \cellcolor{green!15}\bf 84.40  ($\uparrow$2.69\%)\\
  & \method{} w/o  BLS & 88.94 & 82.87 & 75.60 & 72.66 & 83.42 & 89.34 & 93.72 & 83.79 \\
  & \method{} w/o  SPA & 88.21 & 82.59 & 73.83 & 71.33 & 83.42 & 89.35 & 93.51 & 83.18 \\
  \bottomrule
  \end{tabular}
  }
  \label{tab:results}
  \vspace{-2ex}
\end{table}
Table \ref{tab:results} exhibits the performance of {\method} and baselines on seven datasets using PaLM 2-S and  PaLM 2-L as the backbone.
From the results, we have the following findings:
\textbf{First}, we can see that leveraging explanations often improves LLM in-context learning. This enhancement is particularly pronounced when the final prediction is aggregated from multiple predictions sampled from the LLM. Conversely, the standard EP pipeline sometimes even hurts the performance, especially for larger models.
\textbf{Second}, despite its complex design, the latest baseline FLamE often falls short compared to other baselines, which suggests that fine-tuning an additional classifier is particularly important for FLamE and it might be less compatible with the LLM in-context learning framework.
\textbf{Third}, we notice that {\method} can consistently outperform all other methods across both the PaLM 2-S and PaLM 2-L backbones in nearly all datasets, which indicates that {\method} provides a reliable way to improve LLM in-context learning over different tasks.
\textbf{Finally}, When comparing {\method} with its own variants (e.g. w/o BLS and SPA), it's observed that the original \method{} consistently holds an advantage, indicating the necessity of both PW and SA components in maximizing performance.


\subsection{Results on Open-source Models}
\label{sec:open_source}
In order to demonstrate the generalizability of our \method  framework, as well as promote reproducibility, 
we extend our investigations to open-source LLMs including FLAN-UL2~\citep{tay2022ul2}\footnote{Link: \url{ https://github.com/google-research/google-research/tree/master/ul2}.
We only test on StrategyQA dataset since FLAN-UL2 has been fine-tuned on labeled data from other datasets, thus violating the true few-shot setting.} and Llama-2-7b~\citep{touvron2023llama2}. 
Both models have publicly accessible weights\footnote{Link: \url{https://huggingface.co/meta-llama/Llama-2-7b}.}.
As exhibited in Table \ref{tab:opensource}, we observe that these two models generally perform worse than the PaLM 2 model in the main experiments, as they have fewer parameters, and thus may not perform well on these challenging NLU benchmarks. 
Despite this, the experiment results still align with our prior findings, demonstrating that our proposed techniques can consistently yield performance enhancements across these open-source LLMs. 


\begin{table}[tbp]
\renewcommand\arraystretch{0.95}
  \label{tab:datasets}
  \caption{The main experiments results on open-source models, where ``BLS'' stands for bootstrapped LLM scorer and ``SPA'' stands for soft probability aggregation.}
  \resizebox{\linewidth}{!}{
  \begin{tabular}{cccccccccc}
  \toprule
  \bfseries Model ($\rightarrow$) & {\bfseries FLAN-UL2 (20B)}  & \multicolumn{7}{c}{\bfseries Llama-2 (7B)}  \\
  \cmidrule(lr){2-2}   \cmidrule(lr){3-10}  
 \bfseries Dataset ($\rightarrow$) & \bfseries StrategyQA & \bfseries E-SNLI & \bfseries ANLI-R1 & \bfseries ANLI-R2 & \bfseries ANLI-R3 &  \bfseries ECQA &\bfseries StrategyQA & \bfseries OpenbookQA  & \bfseries Avg.  \\
  \midrule
  ICL\scriptsize{~\citep{brown2020language}} & 61.76 & 51.14	& 34.58	 &36.05	 &27.48	 &45.48	 &53.81 &	47.48 & 42.29
  \\
  PE\scriptsize{~\citep{lampinen2022language}} &  73.42 & 54.25	&37.83	 &37.50 &	34.37 &	52.33 &	56.21 &	56.48 & 47.00\\
  EP\scriptsize{~\citep{wei2022chain}} & 75.46 & 56.90	& 35.41	 &39.16 &	36.04 &	54.45	 &57.17	 &44.35 & 46.21\\
  Self-consistency\scriptsize{~\citep{wang2023selfconsistency}} & 76.01  & 58.79	&40.16	&40.16	&36.16	&55.14&	57.12	&60.87 & 49.77\\
  FLamE\scriptsize{~\citep{zhou2023flame}} &  72.17  & 49.32	&36.83&	35.16	&36.50&	45.11	&57.70&	46.23 & 43.84\\ \midrule
  \rowcolor{green!15} \method{} & \bf 78.70 ($\uparrow$ 3.55\%)
  & \bf 60.80 &	\bf 44.50 &\bf	41.66	 &\bf 41.33	 &\bf 60.45 &\bf	59.81	 &64.43 & \bf 53.28 ($\uparrow$ 7.05\%)\\
  \method{} w/o BLS & 77.31 & 59.54&	43.45&	41.33&	40.33&	60.34&	59.62	&\bf 65.06& 52.81 \\
  \method{} w/o SPA & 77.78 & 58.50 &	41.33 &	40.16 &	35.33 &	54.97 &	57.40 &	61.71 & 49.91  \\
  \bottomrule
  \end{tabular}
  }
  \label{tab:opensource}
\end{table}


\begin{table}[!tp]
\renewcommand\arraystretch{0.95}
\centering
  \caption{The study on different scoring approaches. Note that to ensure fair comparison, we do not use \aggregation{} for our method and baselines.}
  \resizebox{0.9\linewidth}{!}{
  \begin{tabular}{lcccccc}
  \toprule
  \bfseries Dataset ($\rightarrow$) & \multicolumn{2}{c}{\bfseries E-SNLI} & \multicolumn{2}{c}{\bfseries OpenbookQA} & \bfseries StrategyQA  \\
    \cmidrule(lr){2-3}   \cmidrule(lr){4-5}   \cmidrule(lr){6-6}
   \bf Model ($\rightarrow$) & PaLM 2-S & PaLM 2-L & PaLM 2-S & PaLM 2-L & FLAN-UL2 \\
  \midrule
  \rowcolor{green!15} {\method}  & \bf 69.82	& \bf 83.68 &	\bf  83.68 & 	93.51 &\bf 	78.70 
\\
  {\method} w/ PE Negative  &68.90 &	83.91	 &83.54	 &\bf 93.93 &	78.06
\\
  LLM Zero-shot Scoring~\scriptsize{\citep{fu2023gptscore}} & 66.84 & 81.77 & 81.38 & 88.50 & 75.15 \\
  LLM Pairwise Scoring~\scriptsize{\citep{qin2023large}} & 69.25	&82.97	&82.97&	93.14	&76.93\\
  Lexical Scoring~\scriptsize{\citep{ye2022unreliability}} & 
 67.72 &	83.54 &	82.66 &	93.72 &	75.34 \\
  NLI Scoring~\scriptsize{\citep{chen2021can}} &   64.87	& 81.89	 &82.21	 & 91.52	 & 76.11
 \\
  \bottomrule
  \end{tabular}
  }
  \vspace{-1ex}
    \label{tab:ensemble}
\end{table}
\vspace{-1ex}
\subsection{Study on Explanation-aware Ensemble}
We perform additional experiments to further understand the benefit of the explanation-aware ensemble, and the result is shown in Table \ref{tab:ensemble}. 

\textbf{Performance w/ Different Scoring Methods} We first compare our LLM-based explanation scorer with a few alternative methods including (1) \emph{lexical scoring}, which estimates the reliability of explanations via the lexical gap~\citep{ye2022unreliability}, and (2) \emph{NLI Scoring} that uses an NLI model to verify the reliability of explanations. 
In this work, we use MT5-XXL~\citep{xue-etal-2021-mt5} fine-tuned on NLI datasets as the scorer. 
Overall, we observe that our model outperforms these models in most of the cases, indicating that LLM has a strong capacity for estimating the quality of the explanations. 
In addition, we observe that pairwise scoring does not perform well for weighting the predictions. This is because it was originally proposed for text ranking tasks, while there are many differences between it and our scenarios, including input formats and relevance signals.

\textbf{Performance w/ Different Bootstrapping Strategies}  To justify the design of leveraging the Explain-then-Predict (EP) pipeline to generate negative demonstrations, we also consider other ways including removing demonstrations as well as using the Predict-then-Explain (PE) pipeline. 
Overall, in many cases, using the EP pipeline leads to better results, as we observe that the PE pipeline sometimes causes the \emph{false negative} issue: it will first generate incorrect predictions but followed with reasonable explanations. 
However, when the model performs reasonably well (e.g. PaLM 2-L on OpenbookQA), then it may make less erroneous prediction during the bootstrapping step, which may lead to insufficient training signals for {\method} to perform well. In addition, no matter whether PE and EP is used, they both largely outperform the baseline where no demonstration is given, necessitating the role of demonstration for explanation-aware ensembling.






\begin{figure}[t!]
    \centering
        \subfigure[E-SNLI, PaLM2-S]{
            \includegraphics[width=0.24\textwidth]{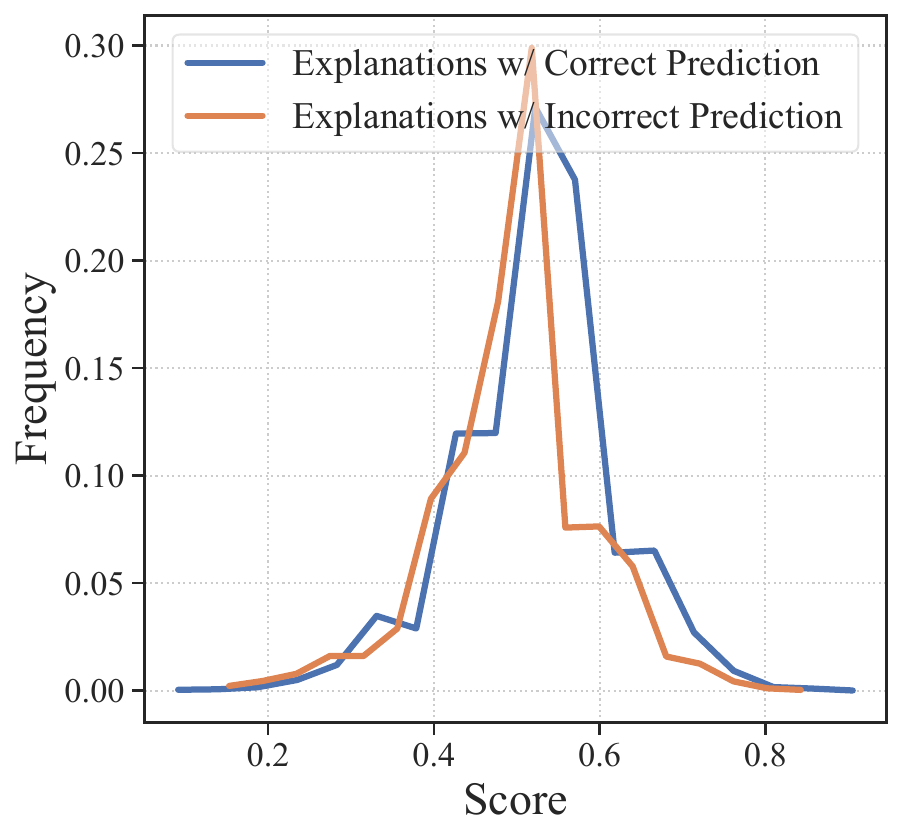}
            \label{fig:temp_esnli}
        } \hspace{-2mm}
        \subfigure[E-SNLI, PaLM2-L]{
            \includegraphics[width=0.24\textwidth]{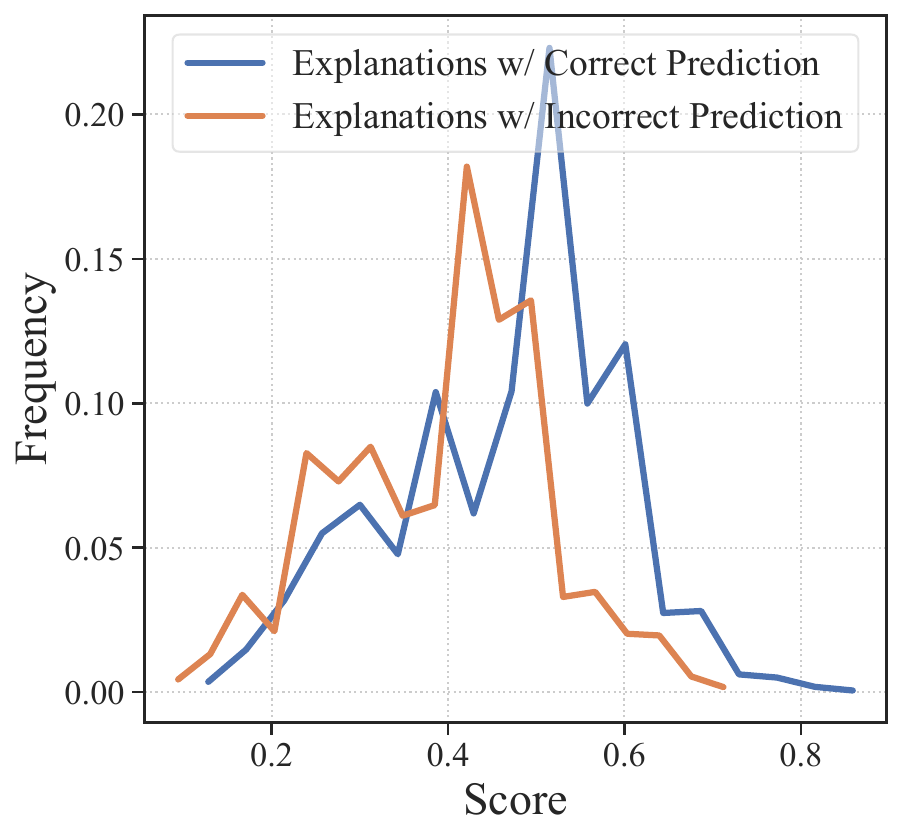}
            \label{fig:temp_openbookqa}
        } \hspace{-2mm}
         \subfigure[OBQA, PaLM2-S]{
            \includegraphics[width=0.24\textwidth]{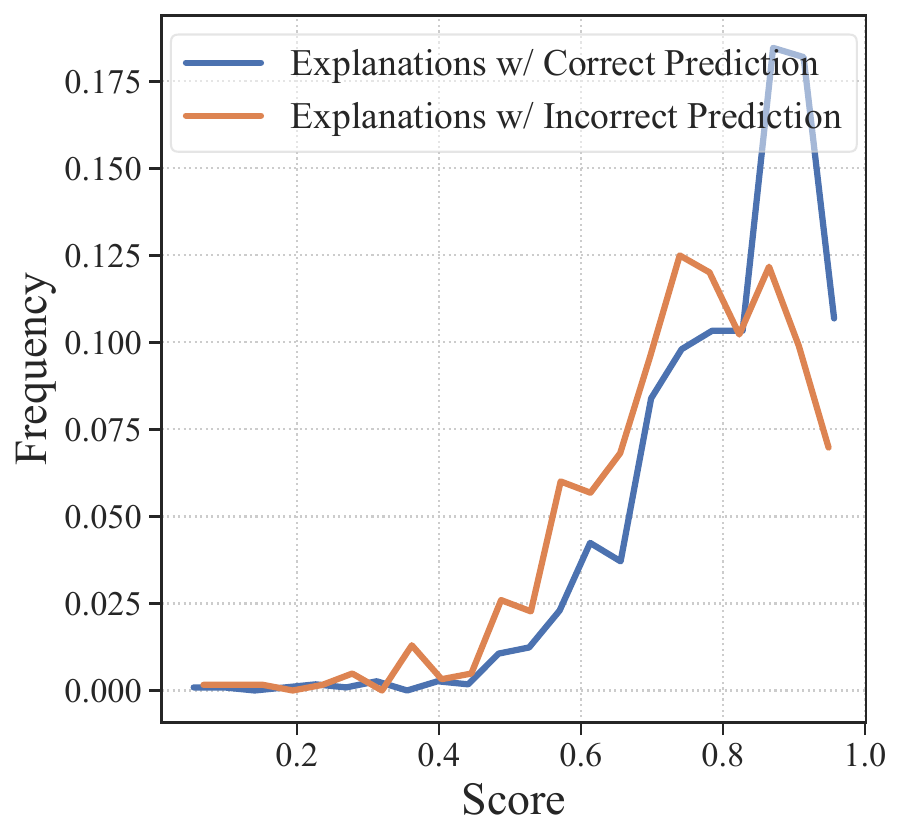}
            \label{fig:obqa_score1}
        } \hspace{-2mm}
         \subfigure[OBQA, PaLM2-L]{
            \includegraphics[width=0.24\textwidth]{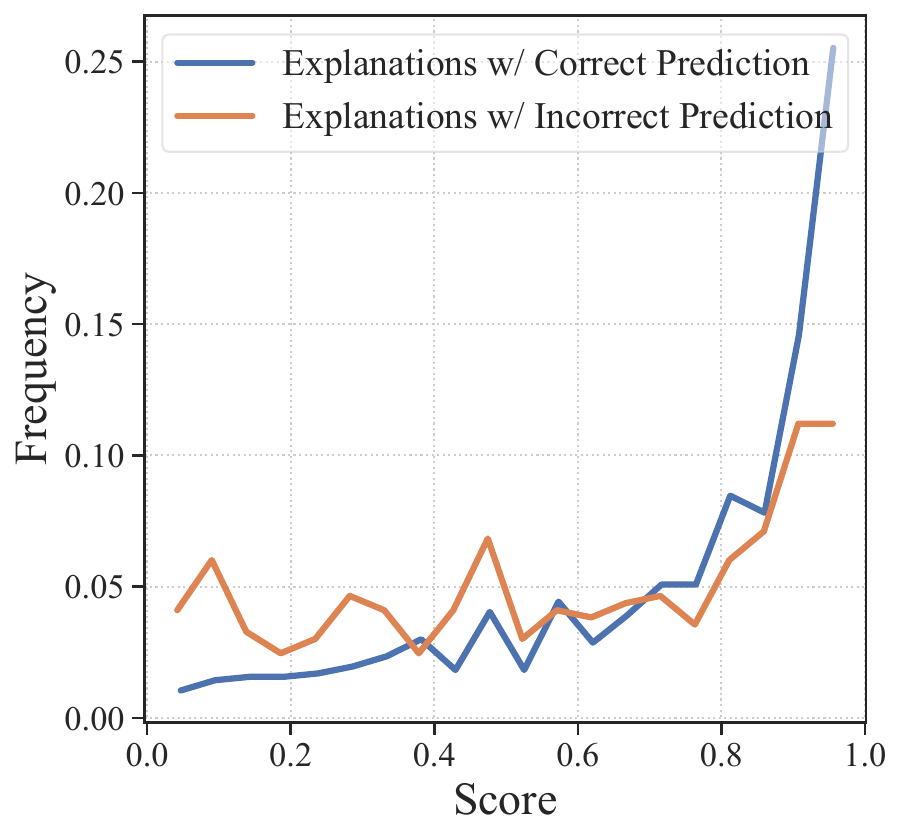}
            \label{fig:obqa_score2}
        } \hspace{-2mm}
        \vspace{-1ex}
        \caption{The score distribution for bootstrapped LLM scorer. OBQA stands for OpenbookQA.\vspace{-2mm}}\label{fig:llm_score}
\end{figure}

\textbf{Score Distribution of Explanations}
To delve deeper into the scores assigned to each explanation and justify that better scores are assigned to explanations with correct answers, we plot the score distribution for explanations with correct predictions\footnote{To eliminate the effect of the sampling randomness, we calculate the prediction based on the soft probability using Eq.~\ref{eq:soft_prob}.} in Figure \ref{fig:llm_score}. Overall, we observe that explanations that lead to correct answers generally have higher scores --- the score distribution is more skewed towards higher values.
Besides, the score distribution using PaLM2-L on explanation with correct and incorrect predictions are more separable, indicating larger models tend to have better scoring performance.

\begin{wrapfigure}[12]{r}{0.3\textwidth}
\vspace{-5.5ex}
\centering
\subfigure{
    \includegraphics[width=0.3\textwidth]{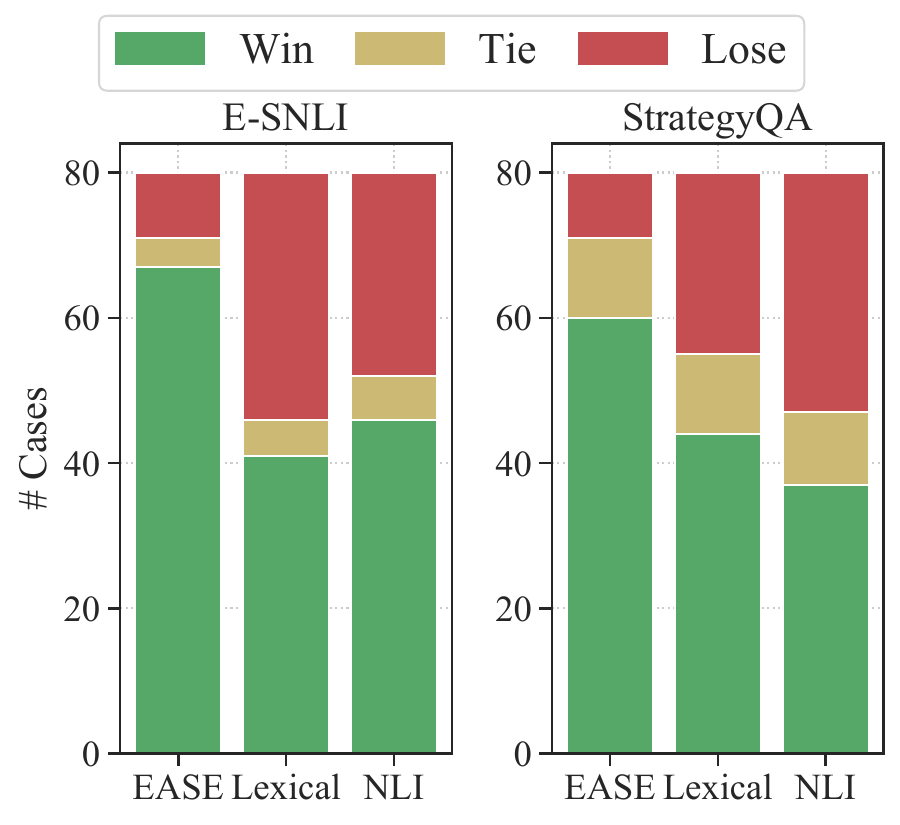}
}
\vspace{-3ex}
\caption{Human Evaluation.}
\vspace{-2ex}
\label{fig:human}
\end{wrapfigure}

\textbf{Human Study on Explanations} 
We conduct additional human studies to further investigate whether the scores generated by LLM are aligned with human preferences.
For each instance, we sample two explanations with \emph{different} predictions as $\{(e_1, p_1), (e_2, p_2)\}$, with one being correct. We compare our approach and two baselines (NLI model, lexical overlap) with human raters: for each pair of explanations, we first ask four humans to determine which explanation is better and use $c_i$ $(i=1,2)$ to denote the number of raters that select $e_i$ as the better one.
Then, we use different models to estimate the score for explanations separately, denoted as $(s_{e_1}, s_{e_2})$. The final judge of ``Win-Tie-Lose'' is determined to be:

\begin{equation}
\small
r = \begin{cases}\textrm{win}, & \text { if } (c_1 > c_2 \text { and }  s_{e_1} > s_{e_2}) \textbf { or }  (c_1 < c_2 \text { and }  s_{e_1} < s_{e_2}) ; \\ 
\textrm{tie}, & c_1=c_2; \\
\textrm{lose}, & \text { if } (c_1 < c_2 \text { and }  s_{e_1} > s_{e_2}) \textbf { or }  (c_1 > c_2 \text { and }  s_{e_1} < s_{e_2}).
\end{cases}  
\end{equation}
On two datasets, we randomly select 80 instances, and the final results are shown in Figure~\ref{fig:human}. 
The cohen's kappa among human raters are 0.75 (E-SNLI) and 0.64 (StrategyQA), which stands for \mquote{substantial agreement}.
Overall, we observe that {\method} aligns with human preferences the best, indicating its better ability to be the proxy for explanation quality estimation. We display more examples on generated explanations and the scores in Appendix~\ref{sec:apd_weight}.

\subsection{Study on Soft Probability Aggregation}
The premise behind \aggregation{} is the potential inaccuracy in the prediction token due to temperature sampling variability.  
To verify this, we calculate the proportion of cases where the prediction token $p_i$ is different than the prediction $p_i \neq \argmax p(\cdot|\cP, x, e_i)$. 

\begin{wraptable}[10]{r}{0.65\linewidth}
  \caption{The study on different probability aggregation approaches. Note that we do not use \weight{} for our method and baselines.}
    \resizebox{0.99\linewidth}{!}{
    \begin{tabular}{lccccc}   \toprule
      \bfseries Dataset ($\rightarrow$) & \multicolumn{2}{c}{\bfseries E-SNLI} & \multicolumn{2}{c}{\bfseries OpenbookQA} & \bfseries StrategyQA  \\
    \cmidrule(lr){2-3}   \cmidrule(lr){4-5}   \cmidrule(lr){6-6} 
   \bf Model ($\rightarrow$) & PaLM 2-S & PaLM 2-L & PaLM 2-S & PaLM 2-L & FLAN-UL2 \\ \midrule
   Inconsistency Ratio & 14.60\% & 10.06\% & 13.96\% & 10.71\% & 10.00\% \\
   \rowcolor{green!15} {\method} &  73.84 & 88.21 & 83.91 & 93.72 & 78.70 \\
   w/ argmax & 73.20 & 87.90 & 83.68 & 93.51 & 78.42 \\
   Cond. Gen \scriptsize{\citep{li-etal-2023-contrastive}} & 70.77 & 82.20 & 78.07 & 84.38 & 72.80 \\  \bottomrule 
    \end{tabular}
    }
    \label{tab:inconsistency}
\end{wraptable}
Overall, as exhibited in Table \ref{tab:inconsistency}, we observe that such inconsistency predictions appear in 10\% to 15\% of the cases, which is not rare in practice. 
By using the soft score, we observe that it will consistently lead to performance boosts. 
The gain is more evident when the inconsistency issue is more severe --- on E-SNLI dataset using PaLM 2-S as the backbone, there exist around 15\% examples with inconsistent predictions. When incorporating {\aggregation{}}, we observe a notable performance gain (from 68.68\% to 73.84\%).  
When compared to other methods for prediction correction, such as using the hard prediction (\ie $\argmax p(\cdot|\cP, x, e_i)$) or generation probability conditioned on different verbalizers, {\method} also achieves better performance.  
More case studies on using soft probabilities are deferred to Appendix~\ref{sec:apd_soft_prob}. 

\subsection{Additional Studies}
\label{sec:addition_experiment}
As {\method} relies on several key components such as prompts and sampling steps, in this section, we study their effect on the final prediction performance, using PaLM 2-S as the backbone model.

\begin{figure}[t!]
    \centering
    \begin{minipage}{0.26\textwidth}
        \centering
        \subfigure{
            \includegraphics[width=\textwidth]{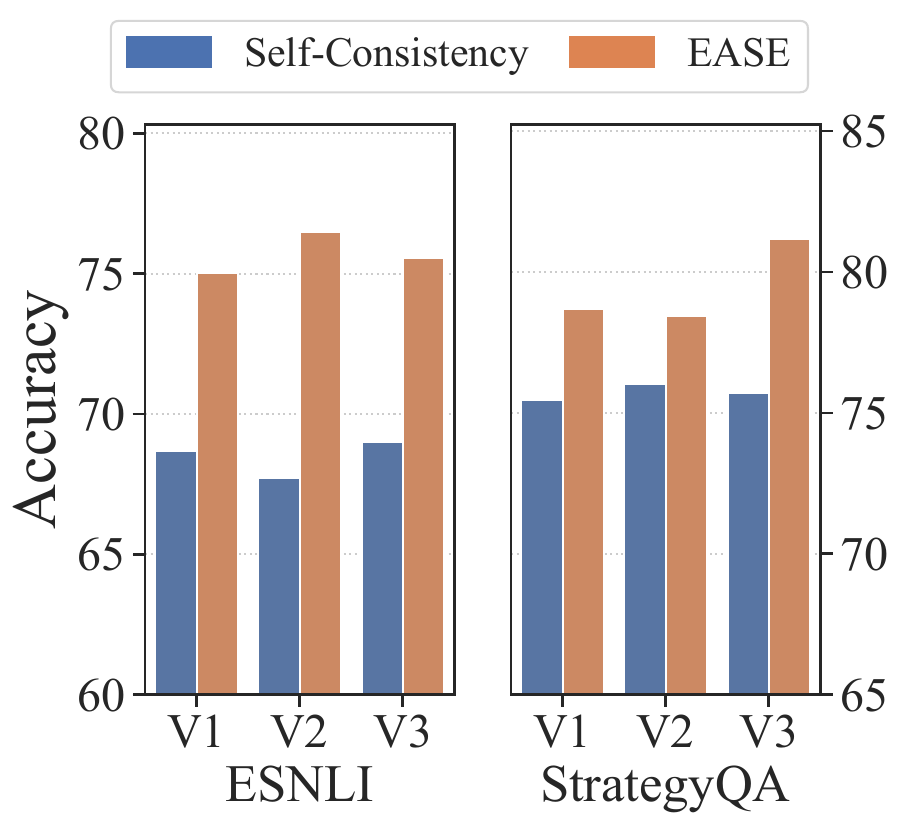}
        }
        \vspace{-2ex}
        \caption{Prompt Format}\label{fig:prompt}
    \end{minipage}%
    \begin{minipage}{0.72\textwidth}
        \centering
        \hspace{-6mm}
        \subfigure[E-SNLI]{
            \includegraphics[width=0.32\textwidth]{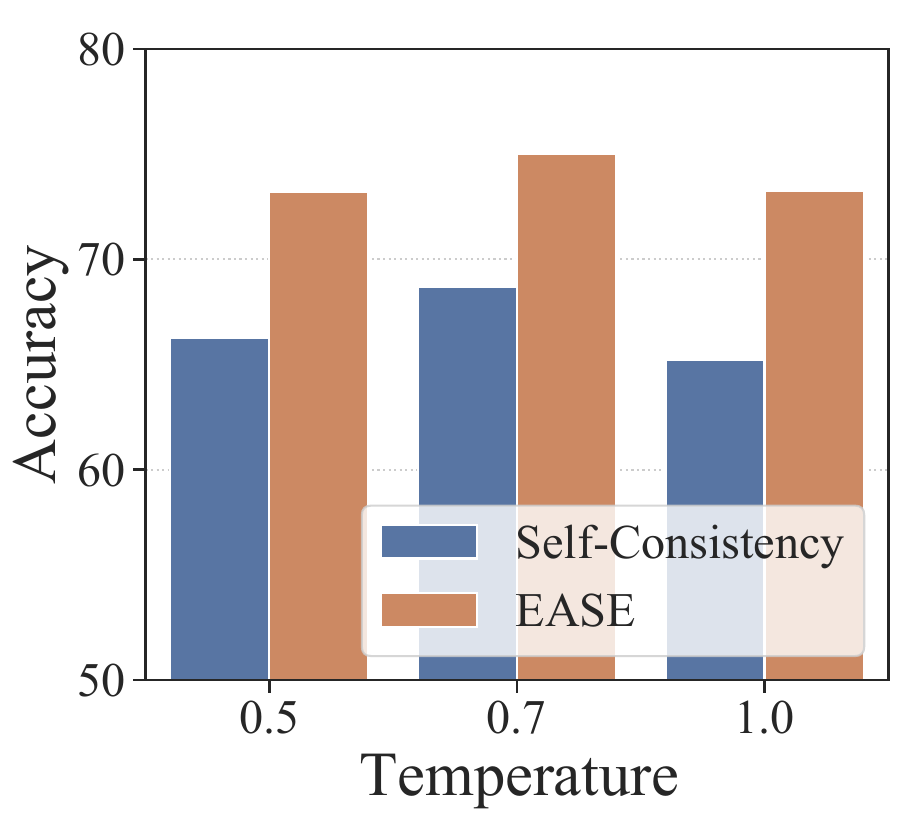}
            \label{fig:temp_esnli}
        } \hspace{-3mm}
        \subfigure[OpenbookQA]{
            \includegraphics[width=0.32\textwidth]{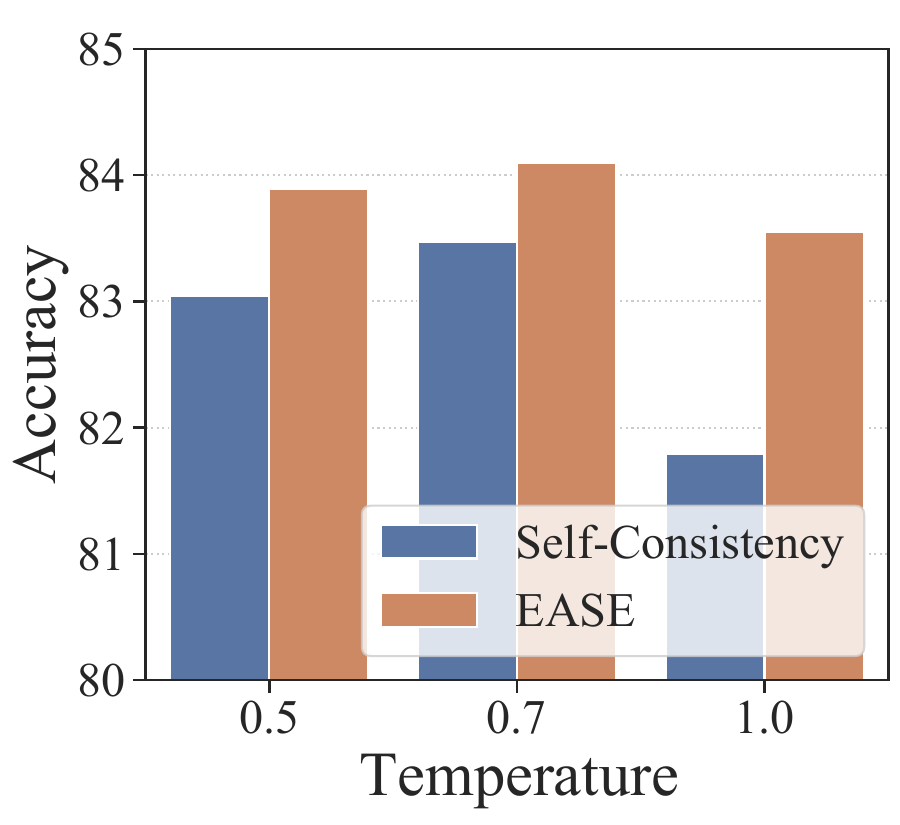}
            \label{fig:temp_openbookqa}
        } \hspace{-3mm}
        \subfigure[StrategyQA]{
            \includegraphics[width=0.32\textwidth]{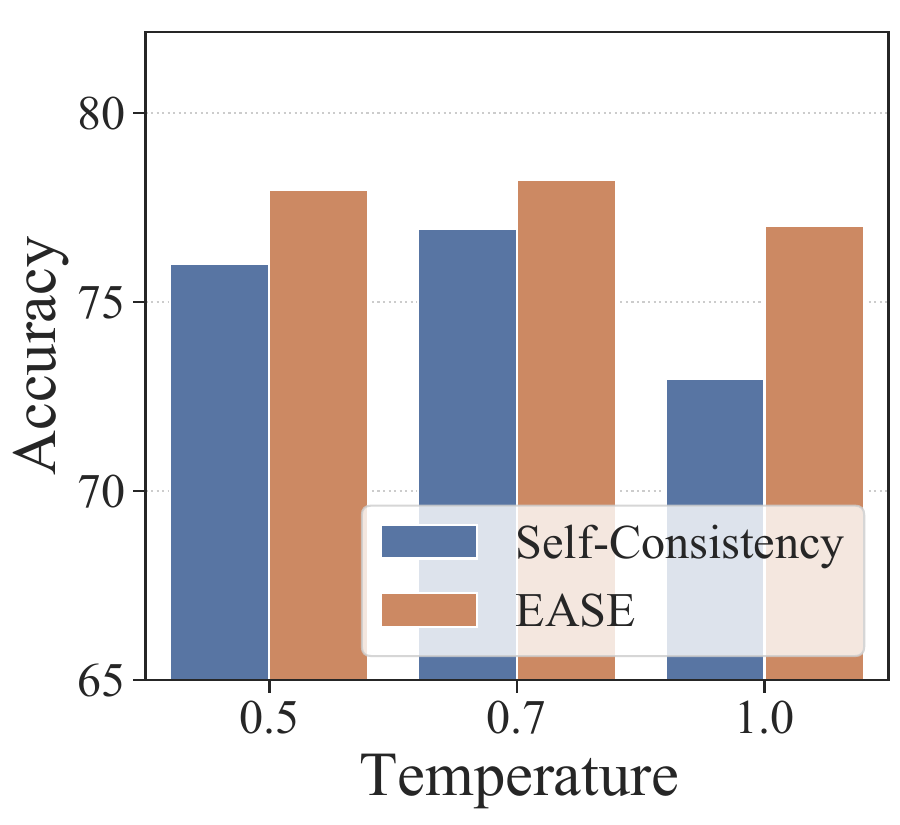}
            \label{fig:temp_strategyqa}
        } \hspace{-6mm}
        \vspace{-2ex}
        \caption{Effect of different temperatures. }\label{fig:temperature}
    \end{minipage}%
    \vspace{-2ex}
\end{figure}

\textbf{Effect of the Sampling Temperatures and Prompt Templates} 
We study the robustness {\method}  to different prompt templates by choosing three different prompt formats from~\citep{bach-etal-2022-promptsource} (the details are in Appendix \ref{sec:apd_addition_prompt}) on two datasets. 
Overall, from Figure \ref{fig:prompt} we observe that {\method} is robust to them as all of the prompt formats lead to performance gains when compared to the strongest baseline self-consistency. 
Similarly, in Figure \ref{fig:temperature}, we observe that {\method} also performs better than baseline under all temperature settings, further justify its robustness across different settings.

\begin{figure}[t!]
    \centering
    \begin{minipage}{0.48\textwidth}
        \centering
        \hspace{-6mm}
        \subfigure[E-SNLI]{
            \includegraphics[width=0.49\textwidth]{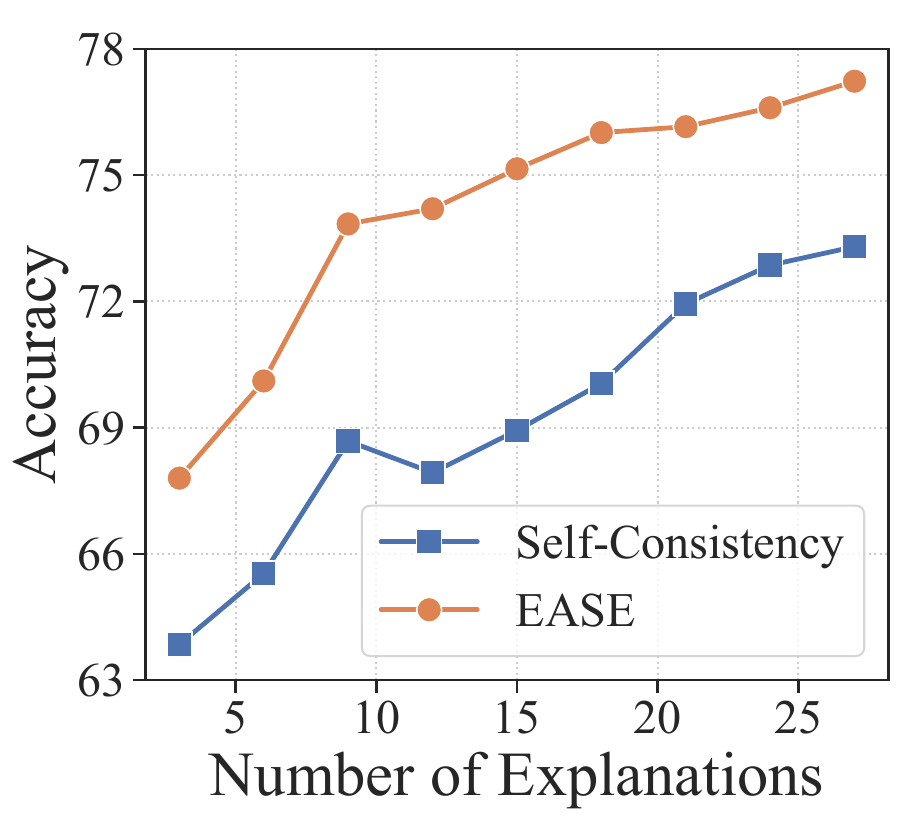}
            \label{fig:num_explanation_esnli}
        } \hspace{-3mm}
        \subfigure[StrategyQA]{
            \includegraphics[width=0.49\textwidth]{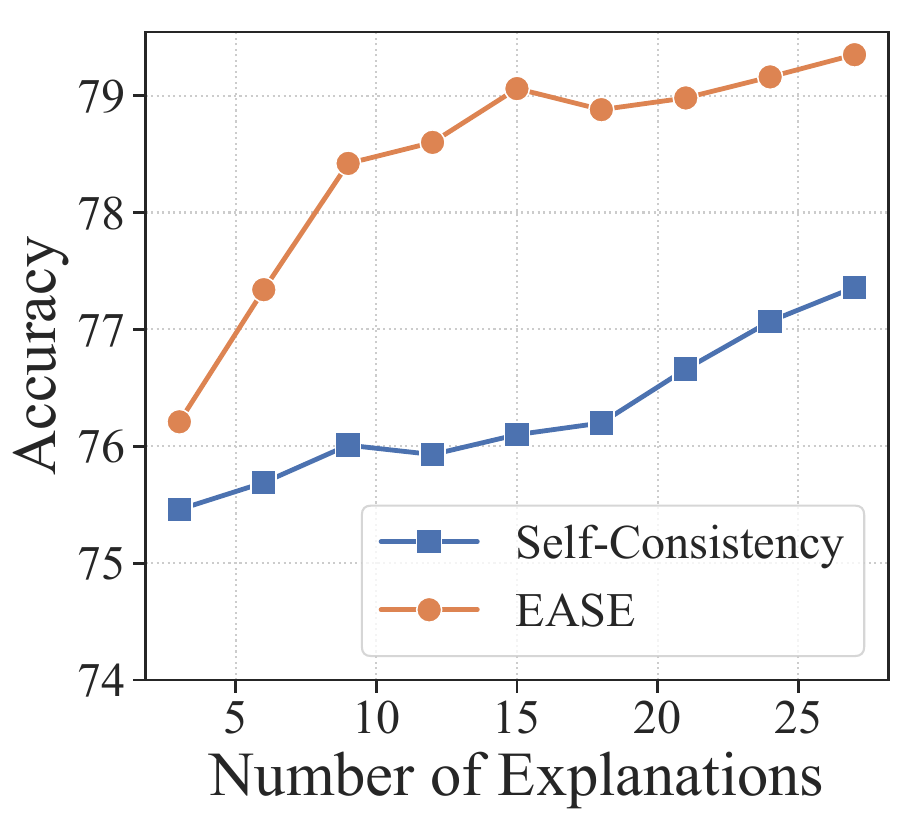}
            \label{fig:num_explanation_strategyqa}
        } 
        \vspace{-2ex}
        \caption{Effect of number of explanations.}\label{fig:num_explanation}
    \end{minipage}%
    \begin{minipage}{0.48\textwidth}
        \centering
        \hspace{-6mm}
        \subfigure[E-SNLI]{
            \includegraphics[width=0.49\textwidth]{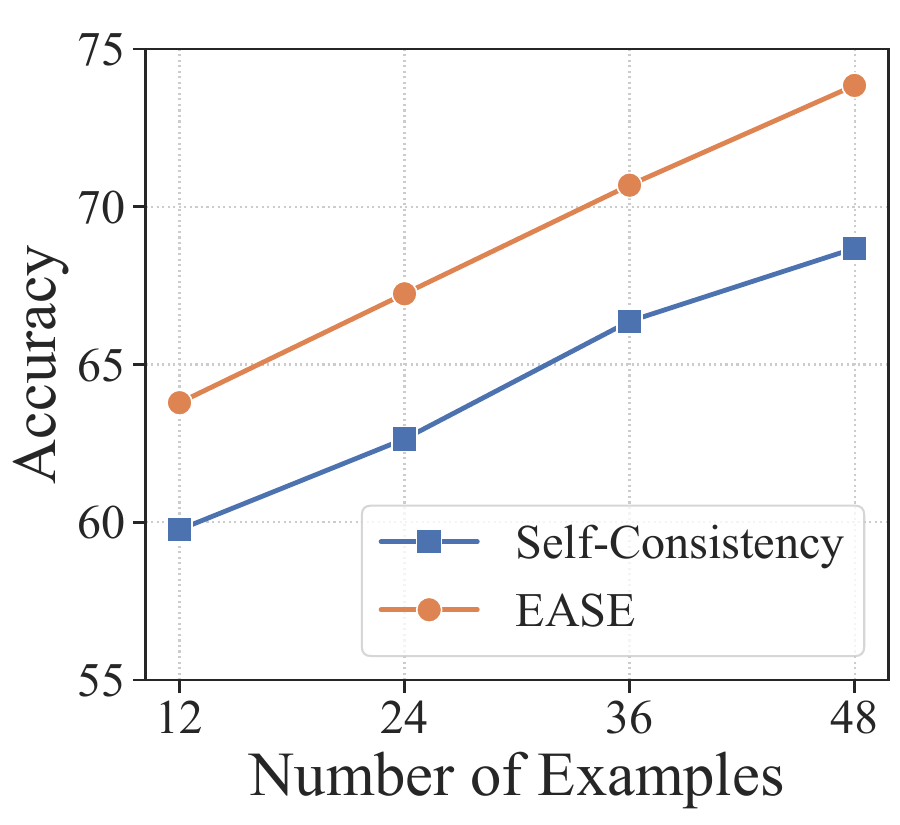}
            \label{fig:num_example_esnli}
        } \hspace{-3mm}
        \subfigure[StrategyQA]{
            \includegraphics[width=0.49\textwidth]{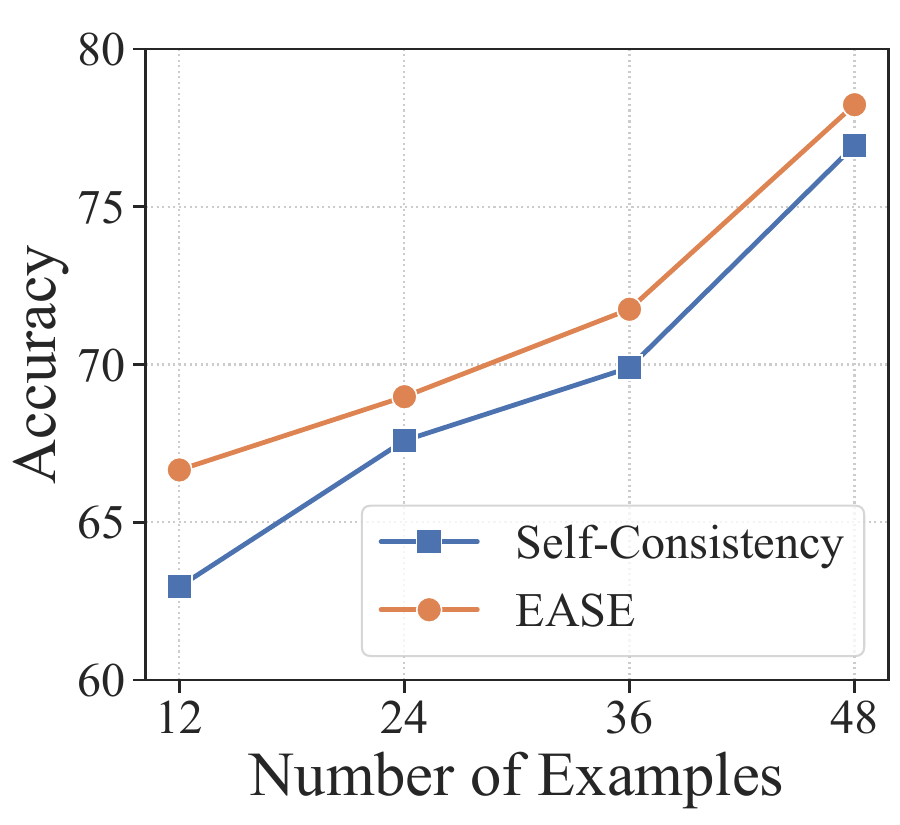}
            \label{fig:num_example_strategyqa}
        }
        \vspace{-2ex}
        \caption{Effect of number of demonstrations.}\label{fig:num_example}
    \end{minipage}%
    \vspace{-2ex}
\end{figure}

\textbf{Effect of the Number of Generated Explanations $\bN$}  
In Figure \ref{fig:num_explanation}, we examine the influence of the number of explanations. On both datasets, increasing the explanations generally improves the performance, while {\method} achieves better performance than the baselines using only 30\% - 40\% of the generated explanations, which can reduce the burden of sampling massive explanations while maintaining the performance.

\textbf{Effect of the Size of demonstrations $\bK$} 
Figure \ref{fig:num_example} illustrates the performance with different size of demonstrations. 
By increasing the number of demonstration $K$,  the performance gradually increases, while {\method} achieves performance gains under all value of $K$. 

\vspace{-1ex}
\section{Conclusion and Discussion}
\vspace{-1ex}
In this work, we empower LLM's in-context learning ability with natural language explanations.
Specifically, we design \weight{} to weight multiple predictions using their associated explanations and realize this idea using a bootstrapped LLM scorer. 
In addition, we leverage a \aggregation{} scheme to mitigate the issue of inconsistent predictions for ensembling. 
We conduct extensive experiments on seven datasets from a diverse task set and show our proposed framework can outperform previous state-of-the-art methods using four LLMs as backbones. 

Notably, while \method~augments in-context learning by weighting predictions through explanations, it does not refine the explanation's content.
For future works, it is potential to leverage techniques such as self-refinement \citep{madaan2023self,ling2023deductive} and debating \citep{du2023improving} to elevate explanation quality and strengthen the model's reasoning abilities.

\section*{Limitations}
In this work, our primary goal is to identify the existing issues to better leverage explanations to empower in-context learning. 
While our approach has shown promise, it also comes with increased computational demands, as both \weight{} and \aggregation{} steps require additional computation overhead. Future work could explore designing more powerful prompts to let LLMs directly output the suffix tokens as quality score~\citep{tian2023just}.
Additionally, our methodology depends on the logits returned in both the \weight{}  and \aggregation{}  processes, making it less suitable to directly adopted black-box LLMs (e.g. ChatGPT,~\citet{openai2023gpt4}).  To approximate the soft score, one strategy is to set the temperature to non-zero value and conduct multiple sampling steps, then use the frequency of the corresponding verbalizers as the proxy of the score. 

Besides, the key assumption of {\method} is that different explanations are of diverse quality, while those explanation leads to correct predictions tend to be of higher quality. 
We mainly conduct empirical experiments to support this point, yet there often exists multiple facets to evaluate the quality of free-text explantions~\citep {chen2023rev,chen2023models,sun2022investigating}. 
More in-depth metrics are needed to faithfully evaluate the quality of free-text explanations and reveal the true inner workings of {\method}.

%



\bibliography{iclr2024,llm}
\bibliographystyle{iclr2024_conference}

\clearpage

\appendix


\section{Datasets Details}

The seven benchmarks in our experiments are all publicly available. Below are the links to downloadable versions of these datasets.
\begin{itemize}
    \item \textbf{E-SNLI}: \url{https://huggingface.co/datasets/esnli}; 
    \item \textbf{ANLI R1/R2/R3}: \url{https://github.com/facebookresearch/anli}; 
    \item \textbf{ECQA}: \url{https://github.com/allenai/feb}; 
    \item \textbf{OpenbookQA}: \url{https://huggingface.co/datasets/openbookqa}; 
    \item \textbf{StrategyQA}: for StrategyQA we use the question-only set from the link \url{https://github.com/google/BIG-bench/blob/main/bigbench/benchmark_tasks/strategyqa}
\end{itemize}

By default, we sample few-shot demonstrations from the train set and sample from the test split for all datasets. 
For OpenbookQA, as the original dataset only contains 500 test examples, in each split we use 100 examples.
For ANLI, as some of the examples contain no explanations, while the explanations for some examples include task-irrelevant information such as `\texttt{I think the computer was confused because so many of the words were similar to the description}'. To reduce the effect of such examples, we remove those examples occurs with term `\texttt{the system}', `\texttt{the computer}', `\texttt{the model}', `\texttt{the AI}', and manually checked all the few-shot demonstrations to ensure that there is no such information in explanations.  

\section{Prompt Formats}
In this section, we list the prompts used in our experiments.
\subsection{Prompt Format For In-context Learning}
In this step, we list the prompt for generating the explanations and predictions. Many of the prompt formats are adapted from \citep{bach-etal-2022-promptsource}. 
Note that the blue text is instance-dependent, while the red text is the model's expected output. 
\subsubsection{E-SNLI}
\clearpage
\begin{lstlisting}[style=mystyle, caption={Prompt Format for E-SNLI dataset, standard in-context learning.}, label=lst:prompt, escapeinside={<@}{@>}]
In this task, given a premise and a hypothesis, your job is to determine whether the hypothesis can be inferred from the premise.


# demonstrations (no more than 48)
Based on the premise: <@\textcolor{blue}{[premise]}@>, can we infer the hypothesis:  <@\textcolor{blue}{[hypothesis]}@> from the premise? Choose among Yes, Maybe, and No. 
Answer: <@\textcolor{blue}{[Answer]}@>

# test examples
Based on the premise: <@\textcolor{blue}{[premise]}@>, can we infer the hypothesis:  <@\textcolor{blue}{[hypothesis]}@> from the premise? Choose among Yes, Maybe, and No. 
Answer: <@\textcolor{red}{[Answer]}@> 
\end{lstlisting}

\begin{lstlisting}[style=mystyle, caption={Prompt Format for E-SNLI dataset, using predict-then-explain pipeline.}, label=lst:prompt, escapeinside={<@}{@>}]
In this task, given a premise and a hypothesis, your job is to determine whether the hypothesis can be inferred from the premise.


# demonstrations (no more than 48)
Based on the premise: <@\textcolor{blue}{[premise]}@>, can we infer the hypothesis:  <@\textcolor{blue}{[hypothesis]}@> from the premise? Choose among Yes, Maybe, and No. 
Answer: <@\textcolor{blue}{[Answer]}@>
Explanation: <@\textcolor{blue}{[Explanation]}@>

# test examples
Based on the premise: <@\textcolor{blue}{[premise]}@>, can we infer the hypothesis:  <@\textcolor{blue}{[hypothesis]}@> from the premise? Choose among Yes, Maybe, and No. 
Answer: <@\textcolor{red}{[Answer]}@>
<@\textcolor{red}{Explanation: [Explanation]}@>
\end{lstlisting}

\begin{lstlisting}[style=mystyle, caption={Prompt Format for E-SNLI dataset, using explain-then-predict pipeline.}, label=lst:prompt, escapeinside={<@}{@>}]
In this task, given a premise and a hypothesis, your job is to determine whether the hypothesis can be inferred from the premise.


# demonstrations (no more than 48)
Based on the premise: <@\textcolor{blue}{[premise]}@>, can we infer the hypothesis:  <@\textcolor{blue}{[hypothesis]}@> from the premise? Choose among Yes, Maybe, and No. 
Answer: <@\textcolor{blue}{[Answer]}@>
Explanation: <@\textcolor{blue}{[Explanation]}@>

# test examples
Based on the premise: <@\textcolor{blue}{[premise]}@>, can we infer the hypothesis:  <@\textcolor{blue}{[hypothesis]}@> from the premise? Choose among Yes, Maybe, and No. 
Explanation: <@\textcolor{red}{[Explanation]}@>
<@\textcolor{red}{Answer: [Answer]}@>
\end{lstlisting}

\subsubsection{ANLI}
\clearpage 

\begin{lstlisting}[style=mystyle, caption={Prompt Format for ANLI dataset, standard in-context learning.}, label=lst:prompt, escapeinside={<@}{@>}]
In this task, given a premise and a hypothesis, your job is to determine whether the hypothesis can be inferred from the premise.


# demonstrations (no more than 48)
Based on the premise: <@\textcolor{blue}{[premise]}@>, can we infer the hypothesis:  <@\textcolor{blue}{[premise]}@> from the premise? Choose among Yes, Maybe, and No. 
Answer: <@\textcolor{blue}{[Answer]}@>

# test examples
Based on the premise: <@\textcolor{blue}{[premise]}@>, can we infer the hypothesis:  <@\textcolor{blue}{[premise]}@> from the premise? Choose among Yes, Maybe, and No. 
Answer: <@\textcolor{red}{[Answer]}@> 
\end{lstlisting}

\begin{lstlisting}[style=mystyle, caption={Prompt Format for ANLI dataset, using predict-then-explain pipeline.}, label=lst:prompt, escapeinside={<@}{@>}]
In this task, given a premise and a hypothesis, your job is to determine whether the hypothesis can be inferred from the premise.

# demonstrations (no more than 48)
<@\textcolor{blue}{[premise]}@>, Based on the previous passage, is it true that <@\textcolor{blue}{[hypothesis]}@>? Choose among Yes, Maybe, and No. 
Answer: <@\textcolor{blue}{[Answer]}@>
Explanation: <@\textcolor{blue}{[Explanation]}@>

# test examples
<@\textcolor{blue}{[premise]}@>, Based on the previous passage, is it true that <@\textcolor{blue}{[hypothesis]}@>? Choose among Yes, Maybe, and No. 
Answer: <@\textcolor{red}{[Answer]}@>
<@\textcolor{red}{Explanation: [Explanation]}@>
\end{lstlisting}

\begin{lstlisting}[style=mystyle, caption={Prompt Format for ANLI dataset, using explain-then-predict pipeline.}, label=lst:prompt, escapeinside={<@}{@>}]
In this task, given a premise and a hypothesis, your job is to determine whether the hypothesis can be inferred from the premise.

# demonstrations (no more than 48)
<@\textcolor{blue}{[premise]}@>, Based on the previous passage, is it true that <@\textcolor{blue}{[hypothesis]}@>? Choose among Yes, Maybe, and No. 
Answer: <@\textcolor{blue}{[Answer]}@>
Explanation: <@\textcolor{blue}{[Explanation]}@>

# test examples
<@\textcolor{blue}{[premise]}@>, Based on the previous passage, is it true that <@\textcolor{blue}{[hypothesis]}@>? Choose among Yes, Maybe, and No. 
Explanation: <@\textcolor{red}{[Explanation]}@>
<@\textcolor{red}{Answer: [Answer]}@>
\end{lstlisting}
\subsubsection{ECQA \& OpenbookQA}
As both ECQA \& OpenbookQA are multi-choice classification tasks, we use the same prompt formats for them. 
\clearpage 
\begin{lstlisting}[style=mystyle, caption={Prompt format for multi-choice QA, standard in-context learning.}, label=lst:prompt, escapeinside={<@}{@>}]
In this task, your job is to first read the question as well as the candidate choices. Then, choose one answer from the choices for the question.

# demonstrations (no more than 48)
Given the following options, what do you think is the correct answer to the question below?
Question:  <@\textcolor{blue}{[question]}@>
Choices: <@\textcolor{blue}{[choices]}@>
Answer: <@\textcolor{blue}{[Answer]}@>

# test examples
Given the following options, what do you think is the correct answer to the question below?
Question:  <@\textcolor{blue}{[question]}@>
Choices: <@\textcolor{blue}{[choices]}@>
Answer: <@\textcolor{red}{[Answer]}@> 
\end{lstlisting}

\begin{lstlisting}[style=mystyle, caption={Prompt format for multi-choice QA, using predict-then-explain pipeline.}, label=lst:prompt, escapeinside={<@}{@>}]
In this task, your job is to first read the question as well as the candidate choices. Then, choose one answer from the choices for the question.

# demonstrations (no more than 48)
Given the following options, what do you think is the correct answer to the question below?
Question:  <@\textcolor{blue}{[question]}@>
Choices: <@\textcolor{blue}{[choices]}@>
Answer: <@\textcolor{blue}{[Answer]}@>
Explanation: <@\textcolor{blue}{[Explanation]}@>

# test examples
Given the following options, what do you think is the correct answer to the question below?
Question:  <@\textcolor{blue}{[question]}@>
Choices: <@\textcolor{blue}{[choices]}@>
Answer: <@\textcolor{red}{[Answer]}@>
<@\textcolor{red}{Explanation: [Explanation]}@>
\end{lstlisting}

\clearpage
\begin{lstlisting}[style=mystyle, caption={Prompt format for multi-choice QA, using explain-then-predict pipeline.}, label=lst:prompt, escapeinside={<@}{@>}]
In this task, your job is to first read the question as well as the candidate choices. Then, choose one answer from the choices for the question.

# demonstrations (no more than 48)
Given the following options, what do you think is the correct answer to the question below?
Question:  <@\textcolor{blue}{[question]}@>
Choices: <@\textcolor{blue}{[choices]}@>
Explanation: <@\textcolor{blue}{[Explanation]}@>
Answer: <@\textcolor{blue}{[Answer]}@>

# test examples
Given the following options, what do you think is the correct answer to the question below?
Question:  <@\textcolor{blue}{[question]}@>
Choices: <@\textcolor{blue}{[choices]}@>
Explanation: <@\textcolor{red}{[Explanation]}@>
<@\textcolor{red}{Answer: [Answer]}@>
\end{lstlisting}

\subsubsection{StrategyQA}
\begin{lstlisting}[style=mystyle, caption={Prompt format for StrategyQA, standard in-context learning.}, label=lst:prompt, escapeinside={<@}{@>}]
In this task, given a question, you need to answer True or False.
# demonstrations (no more than 48)
For the question: '<@\textcolor{blue}{[question]}@>', do you think it is the True or False? 
Answer: <@\textcolor{blue}{[Answer]}@>

# test examples
For the question: '<@\textcolor{blue}{[question]}@>', do you think it is the True or False?  
Answer: <@\textcolor{red}{[Answer]}@> 
\end{lstlisting}

\begin{lstlisting}[style=mystyle, caption={Prompt format for StrategyQA, using predict-then-explain pipeline.}, label=lst:prompt, escapeinside={<@}{@>}]
In this task, given a question, you need to answer True or False.
# demonstrations (no more than 48)
For the question: '<@\textcolor{blue}{[question]}@>', do you think it is the True or False? 
Answer: <@\textcolor{blue}{[Answer]}@>
Explanation: <@\textcolor{blue}{[Explanation]}@>

# test examples
For the question: '<@\textcolor{blue}{[question]}@>', do you think it is the True or False?  
Answer: <@\textcolor{red}{[Answer]}@> 
<@\textcolor{red}{Explanation: [Explanation]}@>
\end{lstlisting}

\clearpage
\begin{lstlisting}[style=mystyle, caption={Prompt format for StrategyQA, using explain-then-predict pipeline.}, label=lst:prompt, escapeinside={<@}{@>}]
In this task, given a question, you need to answer True or False.

# demonstrations (no more than 48)
For the question: '<@\textcolor{blue}{[question]}@>', do you think it is the True or False? 
Explanation: <@\textcolor{blue}{[Explanation]}@>
Answer: <@\textcolor{blue}{[Answer]}@>

# test examples
For the question: '<@\textcolor{blue}{[question]}@>', do you think it is the True or False?  
Explanation: <@\textcolor{red}{[Explanation]}@>
<@\textcolor{red}{Answer: [Answer]}@>
\end{lstlisting}

\subsection{Prompt Format For Explanation-aware Ensemble.}

\begin{lstlisting}[style=mystyle, caption={Prompt format for LLM Scoring. Note that we use the probability of the `Answer' token as the proxy for the quality score.}, label=lst:prompt, escapeinside={<@}{@>}]
In this task, you will be given the input for the [task_name] task, your job is to determine whether the explanation provided is a good one for the given input. Please consider the explanation's coherence, informativeness, and consistency with the prediction to evaluate its quality.

# demonstrations (no more than 48)
For '<@\textcolor{blue}{[task input]}@>', can you determine whether the explanation is a good one for the given <@\textcolor{blue}{[task]}@>?
Explanation: <@\textcolor{blue}{[Explanation]}@>
Answer: <@\textcolor{blue}{[Answer]}@> [Yes or No]


# test examples
For '<@\textcolor{blue}{[task input]}@>', can you determine whether the explanation is a good one for the given <@\textcolor{blue}{[task]}@>?
Explanation: <@\textcolor{blue}{[Explanation]}@>
Answer: <@\textcolor{red}{[Answer]}@>
\end{lstlisting}

\subsection{Additional Prompt Format Used in  Prompt Sensitivity Study}
\label{sec:apd_addition_prompt}
In section \ref{sec:addition_experiment}, we have studied the effect of different prompt templates. Here we list them in the following lists. \clearpage

\begin{lstlisting}[style=mystyle, caption={Prompt Format 2 for E-SNLI dataset}, label=lst:prompt, escapeinside={<@}{@>}]
In this task, given a premise and a hypothesis, your job is to determine whether the hypothesis can be inferred from the premise.

# demonstrations (no more than 48)
Based on <@\textcolor{blue}{[premise]}@>, does it follow that <@\textcolor{blue}{[hypothesis]}@>? Choose among Yes, Maybe, and No. 
Answer: <@\textcolor{blue}{[Answer]}@>
Explanation: <@\textcolor{blue}{[Explanation]}@>

# test examples
Based on <@\textcolor{blue}{[premise]}@>, does it follow that <@\textcolor{blue}{[hypothesis]}@>? Choose among Yes, Maybe, and No. 
Explanation: <@\textcolor{red}{[Explanation]}@>
<@\textcolor{red}{Answer: [Answer]}@>
\end{lstlisting}

\begin{lstlisting}[style=mystyle, caption={Prompt Format 3 for E-SNLI dataset}, label=lst:prompt, escapeinside={<@}{@>}]
In this task, given a premise and a hypothesis, your job is to determine whether the hypothesis can be inferred from the premise.

# demonstrations (no more than 48)
Based on the premise <@\textcolor{blue}{[premise]}@>, can we conclude the hypothesis that <@\textcolor{blue}{[hypothesis]}@>? Choose among Yes, Maybe, and No. 
Answer: <@\textcolor{blue}{[Answer]}@>
Explanation: <@\textcolor{blue}{[Explanation]}@>

# test examples
Based on the premise <@\textcolor{blue}{[premise]}@>, can we conclude the hypothesis that <@\textcolor{blue}{[hypothesis]}@>? Choose among Yes, Maybe, and No. 
Explanation: <@\textcolor{red}{[Explanation]}@>
<@\textcolor{red}{Answer: [Answer]}@>
\end{lstlisting}

\begin{lstlisting}[style=mystyle, caption={Prompt format 2 for StrategyQA, using explain-then-predict pipeline.}, label=lst:prompt, escapeinside={<@}{@>}]
In this task, given a question, you need to answer True or False.

# demonstrations (no more than 48)
Answer the question: '<@\textcolor{blue}{[question]}@>', by True or False. 
Explanation: <@\textcolor{blue}{[Explanation]}@>
Answer: <@\textcolor{blue}{[Answer]}@>

# test examples
Answer the question: '<@\textcolor{blue}{[question]}@>', by True or False. 
Explanation: <@\textcolor{red}{[Explanation]}@>
<@\textcolor{red}{Answer: [Answer]}@>
\end{lstlisting}

\clearpage
\begin{lstlisting}[style=mystyle, caption={Prompt format 3 for StrategyQA, using explain-then-predict pipeline.}, label=lst:prompt, escapeinside={<@}{@>}]
In this task, given a question, you need to answer True or False.

# demonstrations (no more than 48)
EXAM: Answer by True of False.
Question: '<@\textcolor{blue}{[question]}@>'
Explanation: <@\textcolor{blue}{[Explanation]}@>
Answer: <@\textcolor{blue}{[Answer]}@>

# test examples
EXAM: Answer by True of False.
Question: '<@\textcolor{blue}{[question]}@>'
Explanation: <@\textcolor{red}{[Explanation]}@>
<@\textcolor{red}{Answer: [Answer]}@>
\end{lstlisting}

\section{Human Evaluation}
Here we provide the guidelines for human evaluation
\begin{lstlisting}[style=mystyle, caption={Human Evaluation Guideline for E-SNLI dataset.}, label=lst:prompt, escapeinside={<@}{@>}]
For this explanation grading task, given the task input (e.g. the premise and hypothesis for the NLI task and the question for the QA task), ground-truth answer, as well as a pair of explanations from the LLM, you job is to determine which explantion will reach the ground-truth answer for that input. 
For the E-SNLI dataset, your task is to predict if the hypothesis is entailed/neutral/contradicts the premise. 
\end{lstlisting}

\begin{lstlisting}[style=mystyle, caption={Human Evaluation Guideline for StrategyQA dataset.}, label=lst:prompt, escapeinside={<@}{@>}]
For this explanation grading task, given the task input (e.g. the premise and hypothesis for the NLI task and the question for the QA task), ground-truth answer, as well as a pair of explanations from the LLM, you job is to determine which explantion will reach the ground-truth answer for that input. 
For the strategyQA dataset, your task is to answer the question with 'True' or 'False'.
\end{lstlisting}

\section{Studies on  Verbalizers for Bootstrapped LLM Scorer}

We investigate the role of verbalizers for representing the \mquote{positive} and \mquote{negative} explanations. We consider three set of verbalizers, namely V1:\mquote{Yes} and \mquote{No}, V2: \mquote{True} and \mquote{False}, and V3: \mquote{Foo} and \mquote{Jaa} using symbolic tuning~\citep{wei2023symbol}. Using PaLM 2-S as the backbone, we observe that the original \mquote{Yes} and \mquote{No} generally perform better. Symbolic tuning does not work as well as other verbalizers with concrete semantics, indicating it may not be strong enough for the explanation scoring task.

\begin{table}[h]
\centering
\caption{Verbalizer Study for Bootstrapped LLM Scorer, using PaLM 2-S as the backbone.}
\begin{tabular}{c|c|c|c}
\toprule
Template & V1 & V2 & V3 \\ 
\hline
E-SNLI       &  75.01 & 73.75 & 74.12\\
StrategyQA  &  78.40 & 78.23 & 76.75 \\
\bottomrule
\end{tabular}
\end{table}

\section{Additional Case Studies}

\subsection{Case study on \weight{}}
\label{sec:apd_weight}
Table~\ref{tab:case_esnli} and~\ref{tab:case2}  give an example of Explanation-aware Ensemble process on E-SNLI dataset. Take the Table~\ref{tab:case_esnli} as an example, where the original prediction using majority voting is \mquote{Entailment}. 
By leveraging LLM to score each explanation, the LLM is able to reduce the effect of some unreliable explanations: for example, explanation 3 copies a part of the premise without providing additional reasoning,  thus it receives a low score (0.369). 
On the contrary, explanation 4 correctly identifies the key reasoning step, thus receiving a higher score (0.677). After weighted majority voting, we are able to obtain the prediction \mquote{Neutral}, which is the ground-truth answer.




\begin{table*}[t]
\centering
\caption{Case study I for \weight{} on E-SNLI Dataset.}
\vspace{-1ex}
\resizebox{0.98\textwidth}{!}{%
\begin{tabular}{p{16.8cm}|l}
\toprule
 \multicolumn{2}{l}{\textbf{Premise}: A man is working on a computer while two people sit and talk in front.} \\  \multicolumn{2}{l}{\textbf{Hypothesis}: The two people sat and chatted for a bit while the IT guy removed the virus.} \\
 \multicolumn{2}{l}
 {\textbf{Ground-truth Label}: \blue{Neutral}, \quad  \quad \textbf{Majority Voting Prediction}: \red{Entail}.} \\
 \hline 
\textbf{Explanation 1}: The two people cannot be sitting and talking in front while the IT guy removes the virus.\\ 
\textbf{Prediction 1}: \red{Contradict} & Score 1: 0.468\\ \hline 
\textbf{Explanation 2}:  There is no proof that there is an IT Person removed the virus. \\ \textbf{Prediction 2}: \blue{Neutral} & Score 2: 0.562 \\ \hline 
\textbf{Explanation 3}:  The two people were sitting and talking.  \\ 
\textbf{Prediction 3}: \red{Entail} & Score 3: 0.369 \\ \hline 
\textbf{Explanation 4}:  There is no evidence that the man is an IT guy, or that he is removing a virus.  \\ \textbf{Prediction 4}: \blue{Neutral} & Score 4: 0.677 \\ \hline 
\textbf{Explanation 5}:  The two people could be talking and the IT guy could be removing the virus.  \\ \textbf{Prediction 5}: \red{Entail} & Score 5: 0.488 \\ \hline 
\textbf{Explanation 6}:  But it's not given that the IT guy is removing a virus. \\ 
\textbf{Prediction 6}: \blue{Neutral} & Score 6: 0.612 \\ \hline 
\textbf{Explanation 7}:  The man on the computer is the IT guy.  \\ 
\textbf{Prediction 7}: \red{Entail} & Score 7: 0.468 \\ \hline 
\textbf{Explanation 8}:  The people sitting and talking are not related to the man working on the computer.  \\ 
\textbf{Prediction 8}: \red{Contradict} & Score 8: 0.447 \\ \hline 
\textbf{Explanation 9}:  The man sitting at the computer is an IT person and is removing the virus.  \\ \textbf{Prediction 9}: \red{Entail} & Score 9: 0.455 \\ 
\bottomrule
\end{tabular}%
}
\label{tab:case_esnli}
\end{table*}

\begin{table*}[t]
\centering
\caption{Case study II for \weight{}  on StrategyQA Dataset.}
\resizebox{\textwidth}{!}{%
\begin{tabular}{p{16.9cm}|l}
\toprule
 \multicolumn{2}{l}{\textbf{Question}: Is Hermione Granger eligible for the Order of the British Empire?} \\
 \multicolumn{2}{l}
 {\textbf{Ground-truth Label}: \blue{False}, \quad  \quad \textbf{Majority Voting Prediction}: \red{True}.} \\ \hline
\textbf{Explanation 1}: The Order of the British Empire is a British order of chivalry rewarding contributions to the arts and sciences, work with charitable and welfare organizations and public service outside the civil service.  Hermione Granger
is a character in the Harry Potter series by J K. \\ \textbf{Prediction}: \red{True} & Score 1: 0.426\\ \hline 
\textbf{Explanation 2}:  The Order of the British Empire is a British Order of Chivalry. Hermione Granger is a fictional character living in a world of magic and wizards. The Order of the British Empire has never been awarded to a fictional character. \\ \textbf{Prediction}:  \blue{False} & Score 2: 0.655 \\ \hline 
\textbf{Explanation 3}:  The Order of the British Empire was founded in 1917 by King George V. Hermione Granger is a British citizen  To be eligible for the Order of the British Empire, a person must have contributed to the arts, science, medicine, charity, industry or government.  \\ \textbf{Prediction}: \red{True} & Score 3: 0.489 \\ \hline 
\textbf{Explanation 4}:  Hermione Granger is a fictional character from the Harry Potter books, not a British citizen.  \\ \textbf{Prediction}:  \blue{False} & Score 4: 0.678 \\ \hline 
\textbf{Explanation 5}:  Hermione Granger is a fictional character in the Harry Potter books. The Order of the British Empire is a British award that can only be received by a British subject .  \\ \textbf{Prediction}:  \blue{False} & Score 5: 0.650 \\ \hline 
\textbf{Explanation 6}:  Hermione Granger is a fictional character in the Harry Potter series  Order of the British Empire honors people who have made a major contribution to the arts, science, medicine, charity, or industry.  Hermione Granger has contributed to the arts by authoring a book on the history of magic. \\ \textbf{Prediction}: \red{True} & Score 6: 0.406 \\ \hline 
\textbf{Explanation 7}:  Hermione Granger is a British citizen  The Order of the British Empire is an order of chivalry.  \\ \textbf{Prediction}: \red{True} & Score 7: 0.452 \\ \hline 
\textbf{Explanation 8}:  The Order of the British Empire is a knighthood.  Hermione Granger is a Hogwarts student.  \\ \textbf{Prediction}: \red{True} & Score 8: 0.406 \\ \hline 
\textbf{Explanation 9}:  Hermione Granger is a Hogwarts student and does not have British citizenship.  \\ \textbf{Prediction}:  \blue{False} & Score 9: 0.602 \\ 
\bottomrule
\end{tabular}%
}
\label{tab:case2}
\end{table*}

\subsection{Case study on \aggregation{}}
\label{sec:apd_soft_prob}

Table~\ref{tab:soft1} and \ref{tab:soft2} provide examples of Soft Probability Aggregation applied to the ESNLI and StrategyQA Datasets, respectively. 
A notable observation is an occasional mismatch between predicted labels and their respective highest probability scores, indicating the potential for inconsistency in LLM predictions. 
By aggregating the probability score in a soft manner, we are able to rectify the final predictions, justifying the effectiveness of \aggregation{}.

\begin{table*}[t]
\centering
\caption{Case study for Soft Probability Aggregation on ESNLI Dataset. The mismatched prediction is highlighted in \red{red}.}
\resizebox{\textwidth}{!}{%
\begin{tabular}{p{17.2cm}|l}
\toprule
 \multicolumn{2}{l}{\textbf{Premise}: A Seattle firefighter standing in front of his truck.} \\ 
 \multicolumn{2}{l}{\textbf{Hypothesis}: The man is standing in front of the ambulance.} \\ 
  \multicolumn{2}{l}
 {\textbf{Ground-truth Label}: \blue{Contradict}, \quad  \quad \textbf{Majority Voting Prediction}: \red{Neutral}.} \\ \hline
\textbf{Explanation 1}: The man is standing in front of the truck and the man is standing in front of the ambulance. \\ \textbf{Prediction}: Neutral & Score 1: [0.369, 0.419, 0.175]\\ \hline 
\textbf{Explanation 2}:  An ambulance is different than a truck. \\ \textbf{Prediction}: Contradict & Score 2: [0.001, 0.042, 0.953] \\ \hline 
\textbf{Explanation 3}: The Seattle firefighter is standing in front of his truck.  The truck is not necessarily an ambulance.  \\ \textbf{Prediction}: Neutral & Score 3: [0.002, 0.880, 0.105] \\ \hline 
\textbf{Explanation 4}:  Ambulances are not fire trucks.   \\ \textbf{Prediction}: \red{Neutral} & Score 4: [0.001, 0.372, 0.614] \\ \hline 
\textbf{Explanation 5}: The guy is standing in front of a truck.  \\ \textbf{Prediction}: Entail & Score 5: [0.391, 0.194, 0.378] \\ \hline 
\textbf{Explanation 6}: The man is standing in front of the truck. \\ \textbf{Prediction}: Entail & Score 6: [0.418, 0.198, 0.364] \\ \hline 
\textbf{Explanation 7}: Firetrucks and ambulances are different. \\ \textbf{Prediction}: Contradict & Score 7: [0.000, 0.004, 0.995] \\ \hline 
\textbf{Explanation 8}: A Seattle firefighter is a man. \\ \textbf{Prediction}: Entail & Score 8: [0.928, 0.036, 0.027] \\ \hline 
\textbf{Explanation 9}:  The firefighter isn't necessarily an ambulance driver. \\ \textbf{Prediction}: Neutral & Score 9: [0.001, 0.552, 0.456] \\ 
\bottomrule
\end{tabular}%
}
\label{tab:soft1}
\end{table*}

\begin{table*}[t]
\centering
\caption{Case study for Soft Probability Aggregation on StrategyQA Dataset. The mismatched prediction is highlighted in \red{red}.}
\resizebox{\textwidth}{!}{%
\begin{tabular}{p{17.2cm}|l}
\toprule
 \multicolumn{2}{l}{\textbf{Question}: Could every citizen of Samoa send a letter to a unique JPMorgan Chase employee?} \\  
 \multicolumn{2}{l}
 {\textbf{Ground-truth Label}: \blue{True}, \quad  \quad \textbf{Majority Voting Prediction}: \red{False}.} \\ \hline
\textbf{Explanation 1}: Samoa has a population of about 195,000 people  JPMorgan Chase has 260,000 employees. \\ \textbf{Prediction}: \red{True} & Score 1: [0.614, 0.373]\\ \hline 
\textbf{Explanation 2}:  There are 158,000 citizens of Samoa  JPMorgan Chase has 247,000 employees. \\ \textbf{Prediction}: False & Score 2: [0.317, 0.673] \\ \hline 
\textbf{Explanation 3}: There are roughly 275,000 people in Samoa   JPMorgan Chase has roughly 250,000 employees   Not all of JPMorgan Chase's employees are eligible to receive mail at work.  \\ \textbf{Prediction}: False & Score 3: [0.556, 0.443] \\ \hline 
\textbf{Explanation 4}:  Samoa has a population of 197 751. JPMorgan Chase has about 243 000 employees  worldwide.  There are 50 states in the United States   \\ \textbf{Prediction}: \red{False} & Score 4: [0.494, 0.494] \\ \hline 
\textbf{Explanation 5}: There are around 194,000 people living in Samoa  JPMorgan Chase employs around 250,000 people.  \\ \textbf{Prediction}: False & Score 5: [0.526, 0.464] \\ \hline 
\textbf{Explanation 6}:  There are 195,308 citizens of Samoa  JPMorgan Chase has 257,000 employees. \\ \textbf{Prediction}: True & Score 6: [0.180, 0.809] \\ \hline 
\textbf{Explanation 7}: Samoa has a population of around 196,000 people  JPMorgan Chase has more than 250,000 employees.  \\ \textbf{Prediction}: False & Score 7: [0.614, 0.372] \\ \hline 
\textbf{Explanation 8}: The population of Samoa is about 188 000, J P Morgan Chase employs about 250 000 people \\ \textbf{Prediction}: True & Score 8: [0.374, 0.616] \\ \hline 
\textbf{Explanation 9}:  There are 190,000 citizens of Samoa in 2017  JPMorgan Chase has over 250,000 employees. There are between 10-12 000 letters in an average day.  \\ \textbf{Prediction}: \red{False} & Score 9: [0.465, 0.527] \\ 
\bottomrule
\end{tabular}%
}
\label{tab:soft2}
\end{table*}

\end{document}